\documentclass[journal]{IEEEtran}
\usepackage{amsmath,amsfonts}
\usepackage{bm}
\usepackage{algorithmic}
\usepackage{algorithm}
\usepackage{array}
\usepackage[font=normalsize,labelfont=sf,textfont=sf]{subfig}
\usepackage{textcomp}
\usepackage{stfloats}
\usepackage{url}
\usepackage{verbatim}
\usepackage{graphicx}
\usepackage{cite}
\usepackage{amssymb}
\usepackage{multirow}
\usepackage{color}
\usepackage{xcolor}
\usepackage{threeparttable}
\usepackage{adjustbox}
\usepackage[hidelinks]{hyperref}
\usepackage{diagbox}
\usepackage{pifont}
\begin{document}

\title{Emotion Diffusion Classifier with Adaptive Margin Discrepancy Training for Facial Expression Recognition}

\author{Rongkang Dong, Cuixin Yang, Cong Zhang, Yushen Zuo, and Kin-Man Lam~\IEEEmembership{Senior Member,~IEEE,}
\thanks{Rongkang Dong, Cuixin Yang, Yushen Zuo, and Kin-Man Lam are with the Department of Electrical and Electronic Engineering, The Hong Kong Polytechnic University, Kowloon, Hong Kong (e-mail: rongkang97.dong@connect.polyu.hk, cuixin.yang@connect.polyu.hk, zuoyushen12@gmail.com, enkmlam@polyu.edu.hk).}
\thanks{Cong Zhang is with the School of Control Science and Engineering, Shandong University (e-mail: congzhang@sdu.edu.cn)}}

\markboth{Journal of \LaTeX\ Class Files,~Vol.~14, No.~8, August~2021}%
{Shell \MakeLowercase{\textit{et al.}}: A Sample Article Using IEEEtran.cls for IEEE Journals}

\maketitle

\begin{abstract}
Facial Expression Recognition (FER) is essential for human-machine interaction, as it enables machines to interpret human emotions and internal states from facial affective behaviors. Although deep learning has significantly advanced FER performance, most existing deep-learning-based FER methods rely heavily on discriminative classifiers for fast predictions. These models tend to learn shortcuts and are vulnerable to even minor distribution shifts. To address this issue, we adopt a conditional generative diffusion model and introduce the Emotion Diffusion Classifier (EmoDC) for FER, which demonstrates enhanced adversarial robustness. However, retraining EmoDC using standard strategies fails to penalize incorrect categorical descriptions, leading to suboptimal recognition performance. To improve EmoDC, we propose margin-based discrepancy training, which encourages accurate predictions when conditioned on correct categorical descriptions and penalizes predictions conditioned on mismatched ones. This method enforces a minimum margin between noise-prediction errors for correct and incorrect categories, thereby enhancing the model's discriminative capability. Nevertheless, using a fixed margin fails to account for the varying difficulty of noise prediction across different images, limiting its effectiveness. To overcome this limitation, we propose Adaptive Margin Discrepancy Training (AMDiT), which dynamically adjusts the margin for each sample. Extensive experiments show that AMDiT significantly improves the accuracy of EmoDC, achieving improvements of 6.45\%, 16.29\%, 8.03\%, and 4.86\% over the baseline model with standard denoising diffusion training on the RAF-DB basic subset, RAF-DB compound subset, SFEW-2.0 dataset, and AffectNet dataset, respectively, under 100-step evaluations. Additionally, AMDiT-enhanced EmoDC has better generalization and robustness than state-of-the-art discriminative classifiers.
\end{abstract}


\section{Introduction}

\begin{figure}
    \centering
    \includegraphics[width=\linewidth]{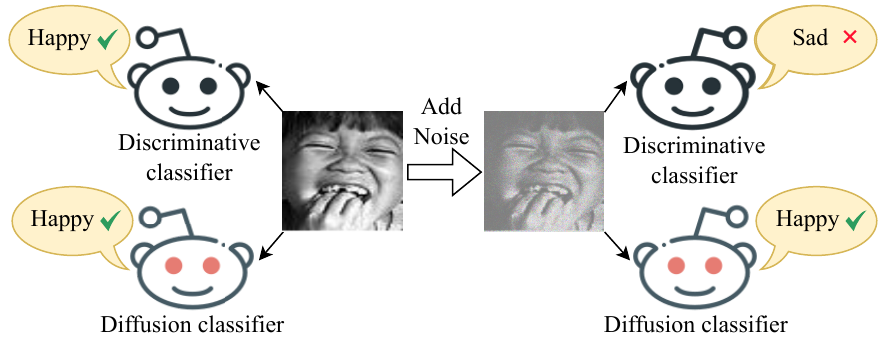}
    \caption{An example comparing classifications of discriminative and diffusion classifiers on clean and noisy images.}
    \label{fig:motivation}
\end{figure}

Facial expressions serve as a primary mechanism for conveying emotional states and are crucial for human non-verbal communication. Automatic facial expression recognition (FER) enables machines to interpret and respond to human emotions during human-computer interactions. The remarkable success of modern deep learning techniques has significantly boosted FER performance \cite{li2022deep, wang2022systematic}. Contemporary deep learning approaches for FER predominantly rely on discriminative classifiers \cite{zhao2023geometry, dong2024bi, dong2024semi, dong2024text, he2016deep}. Given an input sample $\bm{x}$, discriminative classifiers output a categorical probability distribution $p(\bm{c}|\bm{x})$ over facial expression classes using a softmax function, where $\bm{c}$ is the category label. While such models excel at efficient image recognition, they are prone to over-confident predictions \cite{guo2017calibration}, shortcut learning \cite{geirhos2020shortcut, robinson2021can, li2025generative}, and are vulnerable to adversarial attacks \cite{szegedy2014intriguing, goodfellow2015explaining} and perturbations \cite{nguyen2015deep} (an example is shown in Fig. \ref{fig:motivation}). Recently, Li et al. \cite{li2025generative} demonstrated that discriminative classifiers tend to learn shortcuts by over-relying on spurious features rather than core, task-relevant features, leading to failures under even slight distribution shifts.

Despite the predominance of discriminative approaches in FER, the potential of generative classifiers remains largely underexplored, motivating our investigation into their applicability to FER. In contrast to discriminative models, which are prone to shortcut learning, conditional generative models avoid this pitfall by generating complete images, thereby jointly considering both core and spurious features \cite{li2025generative}. As a result, these models inherently reduce reliance on non-essential features. A conditional generative model focuses on modeling the likelihood $p(\bm{x}|\bm{c})$, where $\bm{x}$ and $\bm{c}$ denote the sample and its categorical description. 
For classification tasks, Bayes' theorem is used to compute $p(\bm{c}|\bm{x})$ \cite{ng2001discriminative}: $p(\bm{c}|\bm{x})=p(\bm{x}|\bm{c})p(\bm{c})/p(\bm{x})$. Diffusion models \cite{ho2020denoising, rombach2022high, peebles2023scalable}, a class of state-of-the-art generative models, have demonstrated their powerful generative capacities across a wide range of tasks \cite{rombach2022high, qu2024discriminative, yang2024efficient, zuo2024towards, yang2025vision}. Conditional diffusion models, such as Stable Diffusion \cite{rombach2022high}, incorporate auxiliary text inputs to guide the generation. Recent research has adapted these models into classifiers, known as diffusion classifiers, which have demonstrated strong zero-shot classification capacities \cite{li2023your, clark2023text} and inherent adversarial robustness \cite{chen2024robust, chen2024diffusion}.
The diffusion classifier introduced in \cite{li2023your} assumes a uniform prior $p(\bm{c})$, making the posterior probability proportional to the likelihood:  $p(\bm{c}|\bm{x})\propto p(\bm{x}|\bm{c})$. It computes noise-prediction errors conditioned on categorical descriptions and estimates expectations across timesteps and noises for each category. These expectations are inversely proportional to $p(\bm{x}|\bm{c})$, with the classification result corresponding to the category with the lowest expected error.

Although the diffusion classifier demonstrates impressive zero-shot classification performance, it performs poorly on zero-shot FER because Stable Diffusion is trained on image-text pairs that are not tailored to facial behavior analysis. More importantly, diffusion classifiers require computing noise-prediction errors across all timesteps and classes, resulting in prohibitively high inference times that hinder real-world deployment. Even with efficiency-oriented techniques, such as progressively discarding implausible classes or sharing noise estimate at the same timestep \cite{li2023your, clark2023text}, slow inference remains a persistent challenge. For instance, although a 100-step evaluation improves accuracy, it is nearly 100 times slower than a 1-step evaluation. \textit{These limitations in both accuracy and inference speed highlight the need for further advancements in diffusion-based classifiers for FER.}

A straightforward approach to improving the FER accuracy of diffusion classifiers is to fine-tune them on facial expression datasets, leading to the development of the Emotion Diffusion Classifier (EmoDC). To achieve efficient fine-tuning, low-rank adaptation (LoRA) \cite{hu2022lora} is employed, which significantly reduces the number of trainable parameters. A basic training strategy for EmoDC involves minimizing noise-prediction errors conditioned on correct categorical descriptions (positive prompts) \cite{ho2020denoising, rombach2022high}. However, this objective overlooks the importance of encouraging higher noise-prediction errors for incorrect categorical descriptions (negative prompts), limiting EmoDC's capability to discriminate among different facial expressions. Moreover, a substantial accuracy gap remains between 1-step and multi-step evaluations.
To address these limitations, we introduce discrepancy training, which amplifies noise-prediction errors on negative prompts while minimizing errors on positive prompts for every timestep. To further increase the discriminative capability of EmoDC, we propose margin-based discrepancy training, which enforces a minimum separation between noise-prediction errors of positive and negative prompts. This ensures the model not only distinguishes facial expressions but does so with a clear margin. However, using a fixed margin fails to account for the varying noise-prediction difficulty across different inputs. Small margins should be adopted for inputs where noise prediction is easy (i.e., with low noise-prediction errors), while large margins are needed for those inputs where noise prediction is difficult (i.e., with high noise-prediction errors). To this end, we introduce Adaptive Margin Discrepancy Training (AMDiT), which dynamically adjusts the margin based on sample-wise difficulty. Specifically, AMDiT utilizes the noise-prediction error conditioned on a non-negative description as an adaptive margin, 
allowing it to reflect the varying noise-prediction difficulty across different noisy images. 
Extensive experiments validate the effectiveness of AMDiT in improving FER accuracy and significantly narrowing the performance gap between 1-step and multi-step evaluations. By achieving more accurate recognition at each timestep, AMDiT enables faster inference with fewer timesteps, without substantial degradation in accuracy. Moreover, AMDiT enhances robustness to corrupted facial images, including noise, blur, and adversarial attacks, and consistently outperforms state-of-the-art discriminative classifiers under challenging conditions.

The main contributions of this paper are summarized as follows:
\begin{itemize}
    \item We develop an Emotion Diffusion Classifier (EmoDC) that adopts the generative diffusion model for FER. To the best of our knowledge, this is the first work to explore the diffusion classifier for FER.
    \item We introduce a novel Adaptive Margin Discrepancy Training (AMDiT) method that dynamically adjusts the noise-prediction error margin between positive and negative prompts based on the sample-wise noise-prediction difficulty, effectively enhancing EmoDC's discriminative capability and improving its FER performance.
    \item Extensive experiments demonstrate AMDiT’s effectiveness in boosting accuracy, significantly reducing the accuracy gap between 1-step and multi-step evaluations, and enhancing the robustness of EmoDC against noisy, blurry, adversarially attacked images.
\end{itemize}

The remainder of this paper is organized as follows. Section \ref{sec:relatedwork} reviews related works. Section \ref{sec:method} provides a brief introduction to the preliminaries and presents the proposed AMDiT in detail. Section \ref{sec:experiments} describes the experimental setup and discusses the results. Section \ref{sec:conclusion} gives a summary of this work.

\section{Related Works}
\label{sec:relatedwork}

\subsection{Deep Learning for FER}
Automatic facial affective analysis has been an active research topic for several decades. Recent advancements in deep learning have significantly accelerated the progress in FER, achieving state-of-the-art performance. CNN-based FER approaches employ convolutional neural networks (CNNs) \cite{simonyan2015very, he2016deep, kim2024cvgg, kim2021extensive} to extract spatial features that encode facial expression information \cite{li2017reliable, li2018occlusion}. For example, Kim et al. \cite{kim2024cvgg} proposed the CVGG-19 network, an improved variant of VGG \cite{simonyan2015very}, achieving significantly higher performance while reducing computational cost. In \cite{zhao2023geometry, zhao2022spatial}, human faces were represented using graphs, and FER performance was enhanced by combining CNNs and graph convolutional networks. Attention mechanisms \cite{chen2020stcam, gan2022facial}, including channel, spatial, and temporal attention, have also been adopted to enable models to focus on salient channels, regions, and frames, respectively. Transformer-based FER approaches, such as TransFER \cite{xue2021transfer} and VTFF \cite{ma2021facial}, leverage the Vision Transformer (ViT) \cite{dosovitskiy2021an} architecture to extract global-interactive facial features through the self-attention mechanism. Given the high computational overhead of ViT, APViT \cite{xue2022vision} introduced an attentive pooling mechanism to discard patches and tokens with less useful information. This mechanism effectively reduces computation costs while enabling models to focus on more informative patches and tokens. Multi-modal-based FER approaches utilize multiple modalities, such as speech and text descriptions. Image-text FER models \cite{zhao2023prompting, zhou2024ceprompt, dong2024text} often rely on CLIP \cite{radford2021learning}, which uses natural language supervision for visual feature extraction. Additionally, audio information has been integrated to enhance emotion recognition \cite{bouali2022cross, ryumina2024zero}. While most studies primarily focus on basic facial expressions, humans often display multiple emotions simultaneously through a single expression. Compound FER \cite{du2014compound, dong2024bi, dong2024semi} therefore aims to recognize more complex facial expressions, for example, happy surprise and angry disgust. Despite substantial progress in FER, prior research significantly focuses on discriminative classifiers, leaving generative classifiers largely unexplored. Furthermore, a persistent challenge remains the degraded performance of discriminative FER methods when faced with samples affected by distribution shifts. This work addresses these gaps by investigating the utility of diffusion classifiers for FER, leveraging their generative nature to enhance inference reliability under shifted data distributions.

\subsection{Diffusion Classifiers}
Generative models estimate the data likelihood $p(\bm{x}|\bm{c})$ and can be converted into classifiers to predict the class probability $p(\bm{c}|\bm{x})$ using Naive Bayes \cite{ng2001discriminative}. Generative classifiers are generally more robust than discriminative classifiers \cite{Fetaya2020Understanding, mackowiak2021generative, chen2024diffusion, chen2024robust}. Diffusion models, a type of powerful generative model, have demonstrated remarkable performance across various generative tasks \cite{rombach2022high, qu2024discriminative, yang2024efficient, zuo2024towards}. To enhance their classification capabilities, Mukhopadhyay et al. \cite{mukhopadhyay2023diffusion} extracted features from intermediate layers of the guidance diffusion model \cite{dhariwal2021diffusion} and fed them into classification heads. They compared four types of classification heads evaluated, the attention-based head achieved the best performance. This classification approach essentially mirrors that of discriminative classifiers, where the model outputs the class probability $p(\bm{c}|\bm{x})$. CARD \cite{han2022card} applied a diffusion process to the logit space for both classification and regression tasks. Current conditional diffusion models, like Stable Diffusion \cite{rombach2022high}, generate images conditioned on textual inputs. Due to extensive training on image-text pairs, Stable Diffusion has demonstrated strong zero-shot classification performance on natural images \cite{li2023your, clark2023text}. Jaini et al. \cite{jaini2024intriguing} analyzed the diffusion classifier proposed in \cite{clark2023text} and revealed that its object recognition behavior closely resembles that of humans, particularly in exhibiting shape bias. Robust diffusion classifiers \cite{chen2024diffusion, chen2024robust} were proposed to enhance robustness against adversarial attacks and input perturbations. However, as Stable Diffusion was trained primarily on image-text pairs that are not tailored for facial behavior analysis, its ability to interpret nuanced facial affective behavior is inherently limited, leading to poor zero-shot FER performance. To address this limitation, we incorporate LoRA to efficiently fine-tune the diffusion model on facial expression datasets, equipping the model with domain-specific knowledge for facial expression analysis. To further enhance performance, we introduce a novel AMDiT method, which significantly enhances the EmoDC model by simultaneously improving accuracy, inference speed, and robustness under distribution shifts.

\subsection{Triplet loss}
Triplet loss is a widely used function in machine learning \cite{weinberger2009distance, schroff2015facenet}. It aims to learn an embedding space where the distance between an anchor and a positive example from the same class is minimized, while the distance between the anchor and a negative example from a different class is maximized. FaceNet \cite{schroff2015facenet} was the first to introduce the triplet loss for deep face recognition, along with a triplet selection strategy to mine hard samples within each mini-batch. Wu et al. \cite{wu2017sampling} incorporated a distance-weighted margin to select more informative and stable samples for triplet-loss training. To facilitate faster convergence, several generalizations of triplet-based learning have been proposed. The multi-class \textit{N}-pair loss \cite{rippel2015metric} incorporates multiple negative samples for each anchor, while Magnet loss \cite{sohn2016improved} groups embeddings into class-specific clusters to reduce overlap among local distributions. However, the triplet concept has so far been employed only in recognition tasks that rely on discriminative classifiers. Building upon this concept, we propose a novel training method, AMDiT, for a generative diffusion classifier. AMDiT minimizes noise-prediction errors for positive text descriptions while maximizing those for negative ones, and dynamically adjusts error margins across samples according to their respective noise-prediction difficulties.

\section{Emotion Diffusion Classifier}
\label{sec:method}
In this section, we first provide a preliminary introduction to diffusion models and diffusion classifiers, followed by a detailed description of the proposed AMDiT method.
\subsection{Preliminary}
\subsubsection{Diffusion Models}
Diffusion models \cite{sohl2015deep, ho2020denoising} are generative models that learn to generate data by iteratively adding and removing noise. This involves two main processes: the forward process and the reverse process. Given an observation $\bm{x}_0$ from an unknown real distribution $q(\bm{x})$ and the total number of diffusion steps $T$, the forward process progressively adds Gaussian noise $\epsilon \thicksim \mathcal{N}(0, \bm{I})$ to generate a sequence of noisy samples $\{\bm{x}_{t}\}_{t=1}^{T}$, following a predetermined variance schedule $\{\beta_{t}\}_{t=1}^{T}$. The forward process can be expressed as follows:
\begin{equation}
    q(\bm{x}_{1:T}|\bm{x}_0) = \prod_{t=1}^{T} q(\bm{x}_{t}|\bm{x}_{t-1}),
\end{equation}
where $q(\bm{x}_{t}|\bm{x}_{t-1})=\mathcal{N} \left( \bm{x}_{t};\sqrt{1-\beta_{t}}\bm{x}_{t-1}, \beta_{t}\bm{I} \right)$. A property of the forward process is that a noisy image $\bm{x}_{t}$ at an arbitrary timestep $t$ can be simply obtained as follows:
\begin{equation}
    q(\bm{x}_{t}|\bm{x}_{0}) = \mathcal{N}\left(\bm{x}_{t};\sqrt{\overline{\alpha}_{t}}\bm{x}_{0}, (1-\overline{\alpha}_{t})\bm{I}\right),
\end{equation}
where $\alpha_{t}=1-\beta_{t}$ and $\overline{\alpha}_{t}=\prod_{s=1}^{t}\alpha_{s}$. 

The reverse process gradually removes noises from a Gaussian distribution $p(\bm{x}_T)=\mathcal{N}(\bm{x}_T;0, \bm{I})$ through a learned Gaussian transition. This process is defined as follows:
\begin{equation}
    p_{\theta}(\bm{x}_{0:T}) = p(\bm{x}_{T}) \cdot \prod_{t=1}^{T} p_{\theta}(\bm{x}_{t-1}|\bm{x}_{t}),
\end{equation}
where $p_{\theta}(\bm{x}_{t-1}|\bm{x}_{t})=\mathcal{N}(\bm{x}_{t-1};\bm{\mu}_{\theta}\left(\bm{x}_{t}, t), \bm{\Sigma}_{\theta}(\bm{x}_{t}, t)\right)$. The goal is to train a model to learn the functions $\bm{\mu}_{\theta}(\bm{x}_{t}, t)$ and $\bm{\Sigma}_{\theta}(\bm{x}_{t}, t)$ with parameters $\theta$. Ho et al. \cite{ho2020denoising} derived a denoising diffusion model whose objective is to predict the noise. This model optimizes the approximate evidence lower bound on the log-likelihood as follows:
\begin{equation}
    \log p_\theta(\bm{x}_{0}) \geq  -\mathbb{E}_{\epsilon,t}[w_{t}||\epsilon_\theta(\bm{x}_t, t)-\epsilon||_{2}^{2}] + A 
    \label{eq:vlb},
\end{equation}
where $\epsilon_\theta(\bm{x}_t, t)$ denotes the predicted noise at timestep $t$ by a trainable model, $w_{t}$ is the weight of the noise-prediction error for timestep $t$, and $A$ is a constant. They simplified Equation (\ref{eq:vlb}) by setting $w_{t}=1$ for $t\in[1,1000]$, resulting in:
\begin{equation}
    -\mathbb{E}_{\epsilon,t}[||\epsilon_\theta(\bm{x}_t, t)-\epsilon||_{2}^{2}] + A.
    \label{eq:simplified_loss}
\end{equation}
A standard training objective of diffusion models, denoted as $\mathcal{L}_{\rm base}$, is to minimize the noise-prediction errors, which is written as follows:
\begin{equation}
    \mathcal{L}_{\rm base} = \mathbb{E}_{\bm{x}_{0},\epsilon,t}[||\epsilon_\theta(\bm{x}_t, t)-\epsilon||_{2}^{2}],
    \label{eq:base}
\end{equation}
where the constant $A$ is ignored since it does not depend on the model parameters $\theta$.

\begin{figure}
    \centering
    \includegraphics[width=\linewidth]{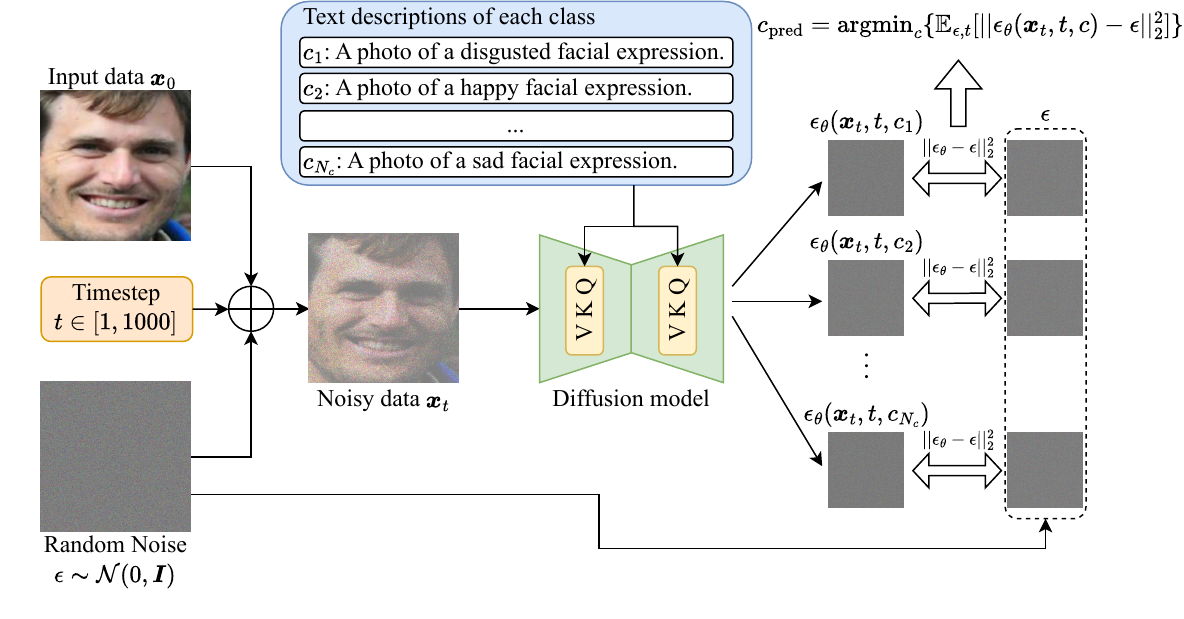}
    \caption{Inference pipeline of the diffusion classifier \cite{li2023your} for FER. $N_{c}$ denotes the total number of categories.}
    \label{fig:inference_pipeline}
\end{figure}

\subsubsection{Diffusion Classifier}
This section provides a brief overview of the diffusion classifier \cite{li2023your}. The diffusion classifier utilizes a conditional diffusion model, such as Stable Diffusion \cite{rombach2022high}, where categorical text descriptions serve as conditional inputs. With this conditional input, Equations (\ref{eq:vlb}) and (\ref{eq:simplified_loss}) are reformulated as follows \cite{li2023your}:
\begin{equation}
    \log p_\theta(\bm{x}_{0}|c) \geq -\mathbb{E}_{\epsilon,t}[||\epsilon_\theta(\bm{x}_t, t, c)-\epsilon||_{2}^{2}] + A, 
    \label{eq:simplifies_loss_conditioned}
\end{equation}
where $c$ denotes the categorical conditions, i.e., text descriptions of the images. 

Fig. \ref{fig:inference_pipeline} illustrates the classification process in diffusion classifiers. The model generates the predicted noises for the input image $\bm{x}_{0}$ at timestep $t$, conditioned on all class descriptions. By evaluating multiple timesteps and noises, the classification result is derived as follows:
\begin{equation}
    c_{\text{pred}}=\text{argmin}_{c}\{\mathbb{E}_{\epsilon,t}[||\epsilon_\theta(\bm{x}_t, t, c)-\epsilon||_{2}^{2}]\}.
    \label{eq:classification}
\end{equation} 

\begin{figure*}[ht]
    \centering
    \includegraphics[width=\textwidth]{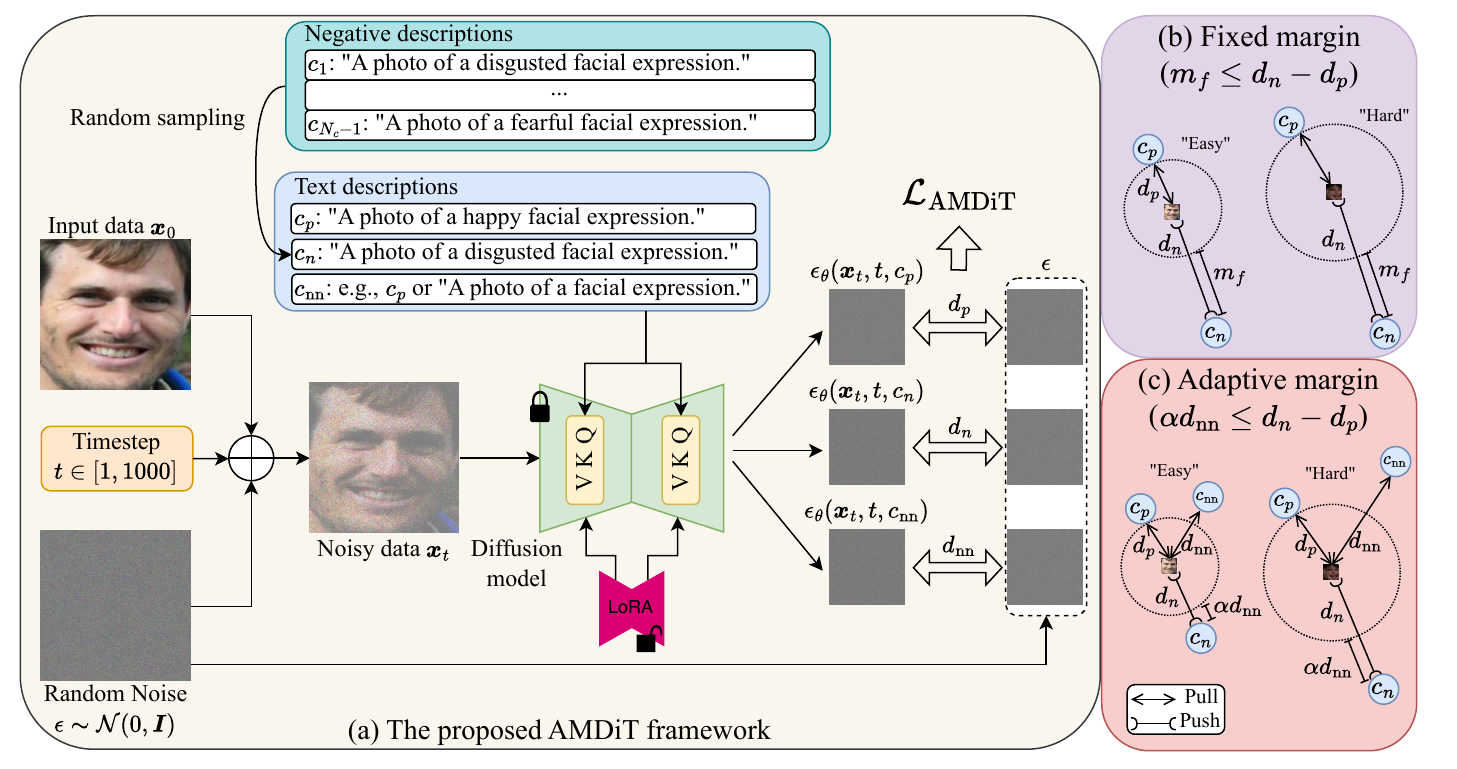}
    \caption{Illustration of: (a) The proposed AMDiT framework, where $d_{*}$ denotes the noise-prediction error $||\epsilon_{\theta}(\bm{x}_{t}, t, c_{*})-\epsilon||_{2}^{2}$ and * represents $p$, $n$, or $\text{\rm nn}$. Specifically, $d_{p}$, $d_{n}$, and $d_{\rm nn}$ represent the noise-prediction errors of a positive image-text pair, a negative image-text pair, and a pair consisting of a facial image and a non-negative text prompt, respectively. These prediction errors are used to compute the $\mathcal{L}_{\rm AMDiT}$ training objective  (Equation~(\ref{eq:amd})). A negative prompt $c_{n}$ for an image is randomly sampled from a pool of negative descriptions (i.e., all categorical descriptions except the positive one). (b) Analysis of training with a fixed margin $m_{f}$. (c) Analysis of training with an adaptive margin $\alpha d_{\rm nn}$. A larger circle radius indicates higher noise-prediction difficulty and therefore a higher error (e.g., a ``hard'' sample), and vice versa.} 
    \label{fig:AMDiT}
\end{figure*}

Previous works \cite{li2023your, clark2023text} have highlighted the exceptional zero-shot classification performance of diffusion classifiers, which typically exhibit low noise-prediction errors for correct class descriptions and high errors for incorrect ones. However, the inference was notably slow because noise predictions must be computed for (a) every class description, (b) every timestep, and (c) numerous sampled noises (as indicated in Equation (\ref{eq:classification})). To improve efficiency, a candidate-class pruning method \cite{li2023your, clark2023text} was adopted, which progressively eliminates unlikely categories by averaging errors over a small number of timesteps. Additionally, shared noise was employed in \cite{clark2023text}, where the same noise sample is evaluated across all classes at a given timestep $t$. In our experiment, we adopt the shared noise method for efficient inference.

\subsection{EmoDC}
Diffusion classifiers have demonstrated strong zero-shot performance on natural images. However, their zero-shot FER performance is poor, achieving an accuracy below 27\% (See Table \ref{tab:incremental_results}). This limitation stems from the insufficient facial affective analysis data in the training process. 

In this study, we employ a diffusion classifier for FER, which we term the Emotion Diffusion Classifier (EmoDC). To equip the model with facial expression knowledge, we fine-tune the diffusion model using LoRA \cite{hu2022lora}, an efficient training method. However, Stable Diffusion \cite{rombach2022high} was trained only on positive image-text pairs, without explicitly amplifying the noise-prediction errors for negative pairs, which limits the model's discriminative power. 

\textbf{Discrepancy training.} To address this issue, we propose increasing the noise-prediction errors for negative image-text pairs. A simple and intuitive approach is to adopt discrepancy training, whose objective function is defined as follows:
\begin{equation}
    \resizebox{0.97\columnwidth}{!}{$\mathcal{L}_{\rm DiT} = \mathbb{E}_{\bm{x}_{0}, t,\epsilon} \left[ ||\epsilon_{\theta}(\bm{x}_{t}, t, c_{p})-\epsilon||_{2}^{2} - ||\epsilon_{\theta}(\bm{x}_{t}, t, c_{n})-\epsilon||_{2}^{2} \right]$} ,
    \label{eq:discrepancyloss}
\end{equation}
where $c_{p}$ and $c_{n}$ denote the positive and negative prompts, respectively. The negative prompts $c_{n}$ are incorrect text descriptions, for example, ``A photo of a disgusted facial expression'' for an image showing a happy facial expression (as illustrated in Fig. \ref{fig:AMDiT}(a)). Rather than computing noise-prediction errors for all negative prompts, we randomly sample one negative prompt per sample, which significantly reduces training time and memory usage. 
However, the objective function in Equation~(\ref{eq:discrepancyloss}) may lead to unstable training: noise-prediction errors for both positive and negative pairs can increase dramatically as training proceeds, even though their discrepancy also grows. This contradicts the goal of diffusion models, to minimize errors for positive descriptions, and significantly degrades the model's capacity to estimate the real data distribution.

\textbf{Margin-based discrepancy training.} Inspired by the triplet loss \cite{schroff2015facenet}, we introduce a margin-based discrepancy training, which enforces a minimal separation between the errors of positive and negative pairs. Formally, it is expressed as follows: 
\begin{align}
    \mathcal{L}_{\text{margin}} = &\mathbb{E}_{\bm{x}_{0}, t,\epsilon} \left[ \max\left\{||\epsilon_{\theta}(\bm{x}_{t}, t, c_{p})-\epsilon||_{2}^{2} - \right.\right. \nonumber \\
    &\left.\left.||\epsilon_{\theta}(\bm{x}_{t}, t, c_{n})-\epsilon||_{2}^{2} +m, 0 \right\} \right],
    \label{eq:fixmargin}
\end{align}
where $m$ denotes the margin.

\textbf{Adaptive Margin Discrepancy Training (AMDiT).} The margin parameter ``$m$'' is typically fixed during training, which fails to account for sample-wise differences in difficulty and thus limits potential performance gains. To overcome this limitation, we propose AMDiT, a novel framework designed to enhance EmoDC's discriminative capability. Fig. \ref{fig:AMDiT}(a) illustrates the proposed AMDiT pipeline. We introduce the use of noise-prediction errors, conditioned on non-negative text prompts $c_{\text{nn}}$, as the adaptive margin. This non-negative description can be the positive prompt $c_{p}$ ($c_{\text{nn}}=c_{p}$), a text description without class information (denoted as $c_{\rm nc}$, e.g., ``A photo of a facial expression."), or a null text ($c_{\text{nn}}=\emptyset$).
The objective function for the proposed AMDiT is formulated as follows:
\begin{align}
    \mathcal{L}_{\text{AMDiT}} = &\mathbb{E}_{\bm{x}_{0}, t,\epsilon} \left[ \max\left\{||\epsilon_{\theta}(\bm{x}_{t}, t, c_{p})-\epsilon||_{2}^{2} - \right.\right. \nonumber \\
        &\left.\left.||\epsilon_{\theta}(\bm{x}_{t}, t, c_{n})-\epsilon||_{2}^{2} + \alpha||\epsilon_{\theta}(\bm{x}_{t}, t, c_{\text{nn}})-\epsilon||_{2}^{2}, 0 \right\} \right],
     \label{eq:amd}
\end{align}
where $\alpha$ controls the weight of the adaptive margin. 

The adoption of this adaptive margin is motivated by two key considerations:

First, a fixed margin lacks adaptability to samples with different noise-prediction difficulties. Figs. \ref{fig:AMDiT}(b) and \ref{fig:AMDiT}(c) compare training under fixed and adaptive margin schemes. The fixed-margin scheme employs a static, predefined margin uniformly across all samples, ignoring variation in noise-prediction difficulty. 
Assigning a large margin to easy samples may impair the model's noise-prediction ability, as it forces excessively large increases in the noise-prediction errors for negative prompts, which can in turn lead to increased noise-prediction errors for positive prompts. Conversely, using a small margin for hard samples has only a limited effect on improving the model's discriminative ability. In contrast, the adaptive-margin scheme dynamically adjusts the margin according to sample-specific difficulty.

\begin{figure*}[t]
    \centering
    \subfloat[]{\includegraphics[width=0.3\linewidth]{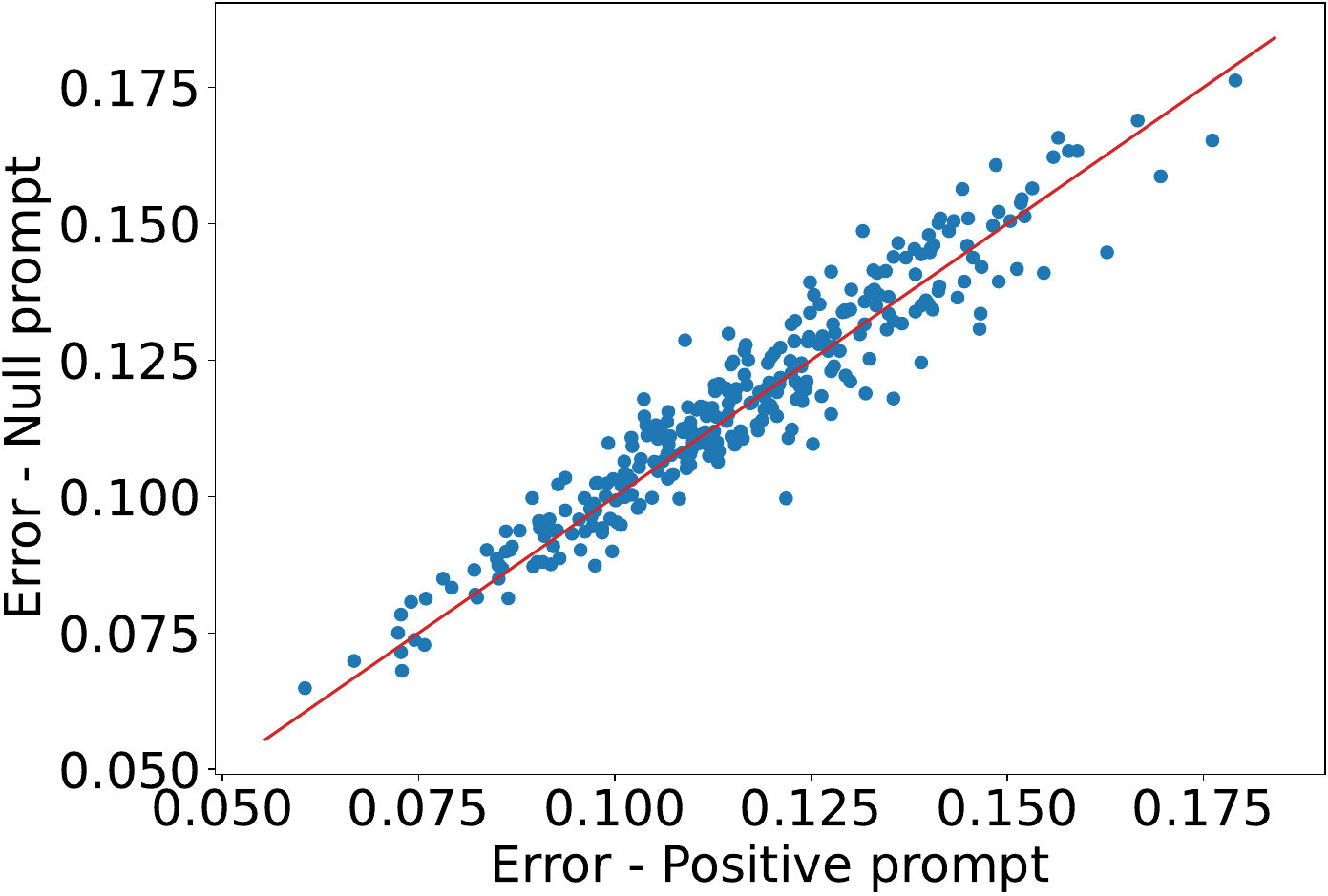}} \hspace{0.02\linewidth}
    \subfloat[]{\includegraphics[width=0.3\linewidth]{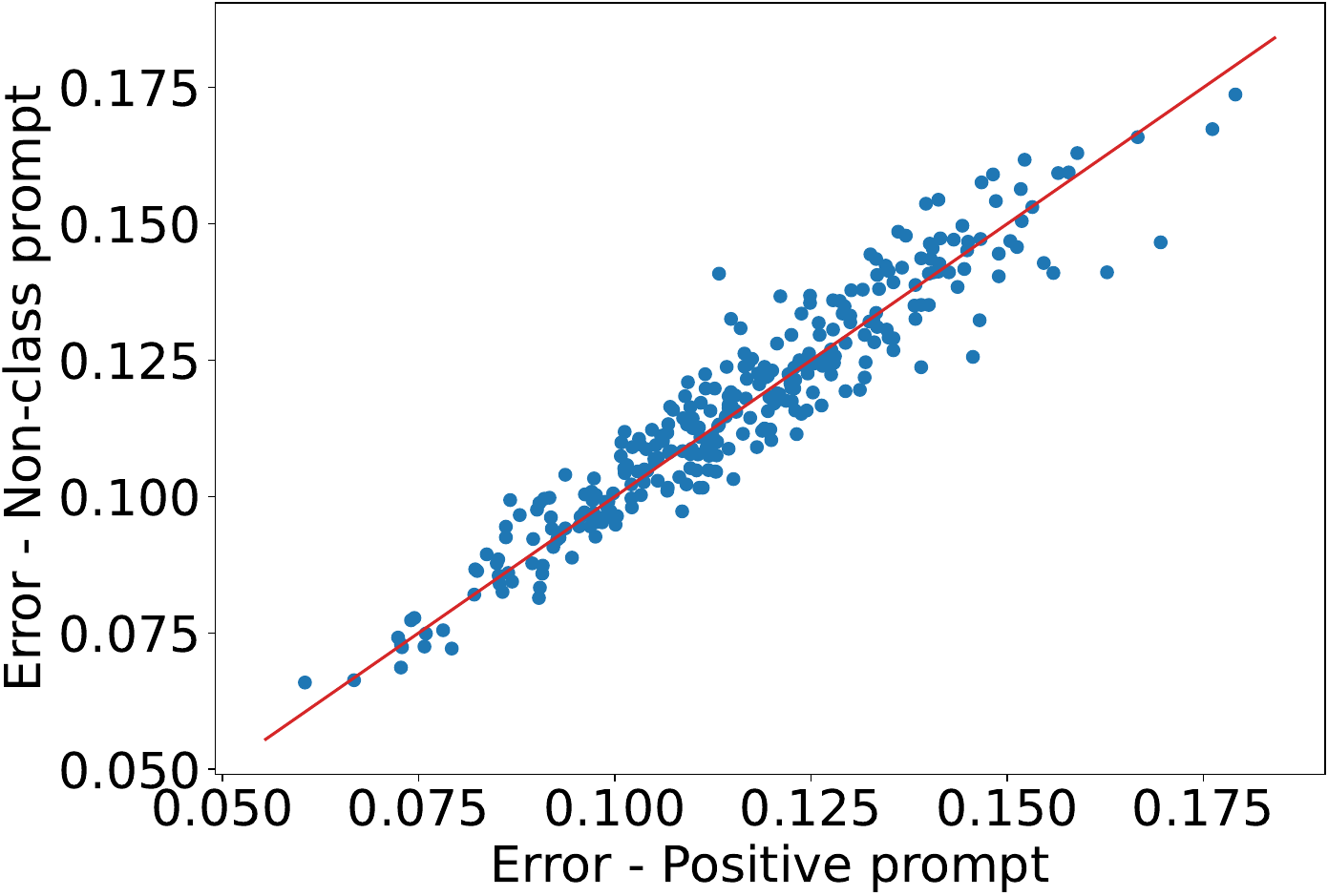}}  \hspace{0.02\linewidth}
    \subfloat[]{\includegraphics[width=0.3\linewidth]{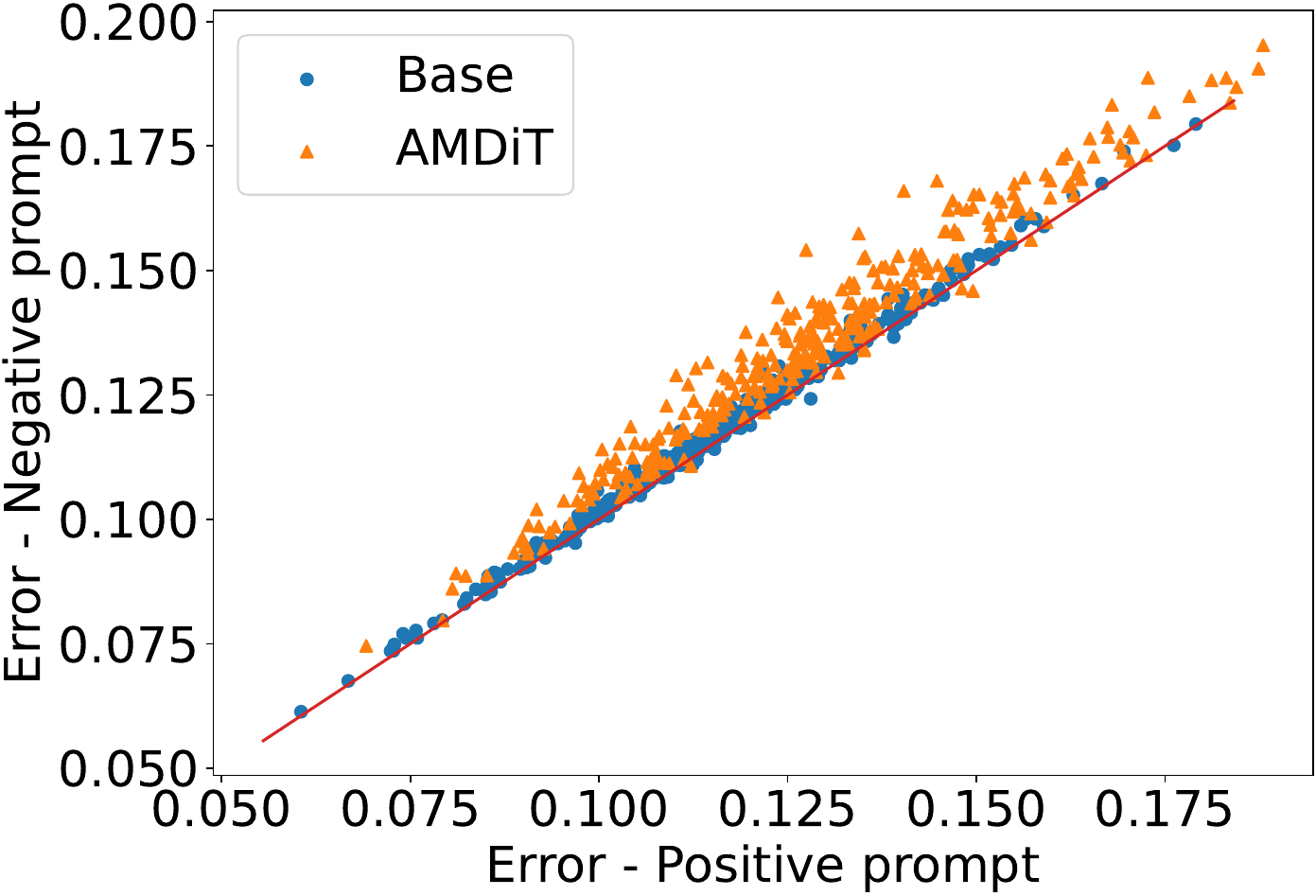}} 
    \caption{Correlations of noise-prediction errors between the positive prompt $c_{p}$ and the null prompt $\emptyset$, non-class prompt $c_{\rm nc}$, and negative prompt $c_{n}$. For each sample, the final error for negative prompts is calculated by averaging the errors across all incorrect categorical descriptions. The red line represents a linear reference with a slope of 1. The ``Base'' model is EmoDC trained using vanilla fine-tuning without negative image-text pairs, with its training objective defined in Equation (\ref{eq:base}). Experiments were conducted on 300 random samples from RAF-DB\_B.}
    \label{fig:noise_linear}
\end{figure*}

Second, denoising diffusion models \cite{ho2020denoising, rombach2022high} are trained to predict noise from noisy images or latent representations. While noise prediction performance varies with the prediction difficulty of different samples, prediction errors show minimal variance when conditioned on different non-negative descriptions for the same image. This observation is supported by Figs. \ref{fig:noise_linear}(a) and \ref{fig:noise_linear}(b), which illustrate the correlation between noise-prediction errors for positive prompts versus null prompts and non-class prompts. Intuitively, noise-prediction error serves as an indicator for noise-prediction difficulty: a smaller error corresponds to lower prediction difficulty, while a larger error signifies greater difficulty. The strong linear correlation with a slope of 1 between non-negative prompts (e.g., null prompt vs. positive prompt) is evident in Figs. \ref{fig:noise_linear}(a) and \ref{fig:noise_linear}(b). This implies that noise-prediction difficulty is primarily determined by the characteristics of the noisy latent (in Stable Diffusion), with minimal influence from the non-negative text prompts. Consequently, noise-prediction errors associated with non-negative prompts can be leveraged as adaptive margins, reflecting sample-specific prediction difficulty. This ensures smaller margins are assigned to easier samples and larger margins to harder ones. In addition, Fig. \ref{fig:noise_linear}(c) depicts the correlation between errors for positive prompts and negative prompts. Notably, most data points lie above the red reference line (with a slope of 1), indicating that errors for negative prompts consistently exceed those for positive prompts. This disparity highlights the effectiveness of adopting diffusion models for FER tasks. Additionally, the proposed EmoDC with AMDiT ($c_{\rm nn}=c_{p}$) exhibits higher errors for negative prompts compared to the EmoDC Base model, suggesting that AMDiT enhances the discriminative capacity of diffusion models in distinguishing facial expressions.

\section{Experiments}
\label{sec:experiments}
\subsection{Datasets}
To demonstrate the effectiveness of the proposed method, we evaluate it on four in-the-wild facial expression datasets, including both basic and compound facial expressions: the RAF-DB basic set (denoted as RAF-DB\_B), the RAF-DB compound set (denoted as RAF-DB\_C) \cite{li2017reliable}, SFEW-2.0 \cite{dhall2011static}, and AffectNet \cite{mollahosseini2017affectnet}. In addition, we also employ FER2013Plus \cite{BarsoumICMI2016} for cross-dataset evaluation.

\textbf{RAF-DB} \cite{li2017reliable} is a large-scale dataset containing nearly 30,000 image samples, divided into two subsets: a basic set and a compound set. Most prior studies have focused on the basic set. The basic set contains 12,271 training images and 3,068 testing images, categorized into seven classes: one neutral facial expression (Ne) and six basic facial expressions, i.e., happiness (Ha), anger (An), sadness (Sa), disgust (Di), surprise (Su), and fear (Fe). The compound set, on the other hand, contains 3,162 training samples and 792 testing samples, comprising eleven compound facial expressions: HaDi (i.e., Happy Disgust), HaSu, SaFe, FeAn, FeSu, SaAn, SaSu, SaDi, AnSu, AnDi, and DiSu. These two subsets are adopted to demonstrate the effectiveness of the proposed method in analyzing both simple and complex facial expressions.

\textbf{SFEW-2.0} \cite{dhall2011static} is a challenging FER benchmark, comprising over 1,500 facial images extracted from movies. The dataset is divided into three sets: a training set with 958 samples, a validation set with 436 samples, and a test set with 372 samples. We follow previous studies \cite{wang2020region, zhao2023geometry, xue2022vision} that use the validation set for evaluation. SFEW 2.0 includes seven categories, identical to those in RAF-DB\_B. 

\textbf{AffectNet} \cite{mollahosseini2017affectnet} is a large-scale FER database, containing more than 290,000 facial expression images. In this study, we employ the seven classes of standard facial expressions. The dataset is divided into two subsets: a training set with 283,900 samples and a validation set with 3,500 samples. The validation set is balanced, with an equal number of samples for each class.

\textbf{FER2013Plus} \cite{BarsoumICMI2016} is a re-annotated version of FER2013 \cite{goodfellow2013challenges}. The dataset is divided into three subsets: a training set with 28,558 samples, a validation set with 3,579 samples, and a test set with 3,573 samples. We adopt the seven basic expressions from the test set for cross-dataset evaluations.

\subsection{Implementation Details}
We employ Stable Diffusion v2.1 \cite{rombach2022high} as the conditional diffusion model to evaluate the performance of the proposed training methods. For efficient training, we adopt LoRA with a rank of 16. The AdamW \cite{loshchilov2018decoupled} optimizer is employed with a learning rate of 0.0001 and a weight decay of 0.01. Gradients are clipped to a maximum value of 1, with a gradient accumulation step of 1. The number of training steps is set to 30,000 for RAF-DB\_B, 10,000 for RAF-DB\_C, and 2,000 for SFEW 2.0. Experiments are conducted on RTX4090 GPUs with a total batch size of 64. The model is trained on images with a resolution of $256\times256$. Images are aligned and cropped to a resolution of $512\times512$ using MTCNN \cite{zhang2016joint}, then resized to $256\times256$. Random horizontal flipping is applied as the only data augmentation technique. 

\subsection{Experimental Results}

\begin{table*}
    \centering
    \caption{Ablation study of the proposed AMDiT method. We report accuracy (\%) of different training methods for Stable Diffusion under 100-step evaluations. DiT means discrepancy training. CoDiT and FMDiT refer to Contrastive noise Discrepancy Training and Fixed-Margin Discrepancy Training, respectively.}
    \label{tab:incremental_results}
    \begin{tabular}{l|ccc|cccc}
    \hline
    Method \textbackslash\ Dataset & DiT & Margin-based & Adaptive margin & RAF-DB\_B  & RAF-DB\_C & SFEW-2.0 & AffectNet \\ \hline
    Zero-shot                      &  \ding{55} & \ding{55} & \ding{55} & 26.34      & 17.68     & 24.31 & 28.14   \\
    EmoDC Base                     &  \ding{55} & \ding{55} & \ding{55} & 83.67      & 51.89     & 48.39 & 59.63   \\
    EmoDC w/ CoDiT                           &  \ding{51} & \ding{55} & \ding{55} & 85.17      & 53.54     & 49.77 & 60.26 \\ 
    EmoDC w/ FMDiT                           &  \ding{51} & \ding{51} & \ding{55} & 87.22      & 63.38     & 52.98 & 61.63  \\
    EmoDC w/ AMDiT ($c_{\text{nn}}=c_{\text{nc}}$) & \ding{51} & \ding{51} & \ding{51}  & 89.08  & 63.51 &\textbf{56.42} & \textbf{64.49} \\ 
    EmoDC w/ AMDiT ($c_{\text{nn}}=c_{p}$)  & \ding{51} & \ding{51} & \ding{51} & \textbf{90.12}  & \textbf{68.18} & 56.19 & 63.89 \\ \hline  
    \end{tabular}
\end{table*}

\begin{table}[t]
    \caption{Accuracy (\%) on RAF-DB\_B with different $\alpha$ values for EmoDC trained with AMDiT.}
    \label{tab:hyperparameter}
    \centering
    \begin{tabular}{c|c|c|c|c}
    \hline
    $\alpha$ & 1-step & 10-step & 100-step & Avg. \\ \hline
    0.005      & 88.72  & 89.28   & 89.41    & 89.14 \\
    0.01       & 89.18  & 89.83   & 90.12    & \textbf{89.71} \\
    0.05       & 88.23  & 88.66   & 88.62    & 88.50 \\
    \hline
    \end{tabular}
\end{table}

\subsubsection{Hyperparameter Sensitivity}
We investigate the impact of different values of $\alpha$ in AMDiT. Evaluations are conducted using different numbers of timesteps, including 1-step evaluation, and uniformly spaced 10-step and 100-step evaluations. Final performance is assessed by averaging accuracies across the 1-step, 10-step, and 100-step evaluations. For the 1-step evaluation, the 100th timestep is used (see analysis in Section \ref{subsubsec:eval_timestep}). In this experiment, we adopt $c_{\text{nn}}=c_{p}$ for AMDiT. The results in Table \ref{tab:hyperparameter} show that model performance varies with different hyperparameter values. A large $\alpha$ may lead to unstable training and suboptimal accuracy, whereas a small $\alpha$ weakens EmoDC's discriminative capability due to insufficient separation constraint. The optimal value is empirically determined to be $\alpha=0.01$, which is used in all subsequent experiments.

\subsubsection{Non-negative Prompt Analysis}
AMDiT uses an adaptive margin derived from the noise-prediction error conditioned on a non-negative prompt, which reflects the noise-prediction difficulty of each sample. To analyze the effect of different non-negative prompts $c_{\text{nn}}$, we conduct experiments on RAF-DB\_B using: (i) $c_{\text{nn}}=$``A photo of a facial expression.'' (class-agnostic prompt $c_{\text{nc}}$), (ii) $c_{\text{nn}}=c_{p}$, and (iii) $c_{\text{nn}}=\emptyset$. The results of the 100-step evaluation are used for this experiment and all subsequent ones, except those in Section \ref{subsubsec:eval_timestep}.
As shown in Table~\ref{tab:non-negativeprompt}, among the three adaptive-margin variants, AMDiT with $c_{\text{nn}}=c_{p}$ achieves the best performance. In addition, all three variants outperform the EmoDC baselines (see Table~\ref{tab:incremental_results}), demonstrating the effectiveness of the proposed AMDiT.

\begin{table}[t]
    \caption{Accuracy (\%) of AMDiT with different non-negative prompts on RAF-DB\_B. Here, the non-class prompt $c_{\text{nc}}=$``A photo of a facial expression.''}
    \label{tab:non-negativeprompt}
    \centering
    \begin{tabular}{c|c|c|c}
    \hline
    $c_{\text{nn}}=$ & $c_{\text{nc}}$ & $c_{p}$ & $\emptyset$ \\ \hline
    Acc.     & 89.08 & \textbf{90.12} & 87.74 \\ \hline
    \end{tabular}
\end{table}

\begin{figure*}
    \centering
    \captionsetup[subfloat]{labelformat=empty}
    \subfloat[RAF-DB\_B]{
        \subfloat{\includegraphics[width=0.24\linewidth]{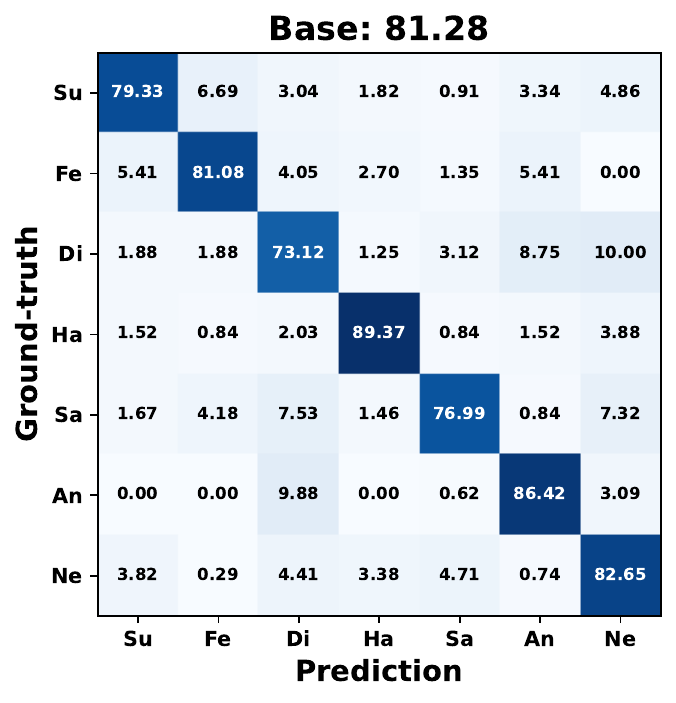}}
        \subfloat{\includegraphics[width=0.24\linewidth]{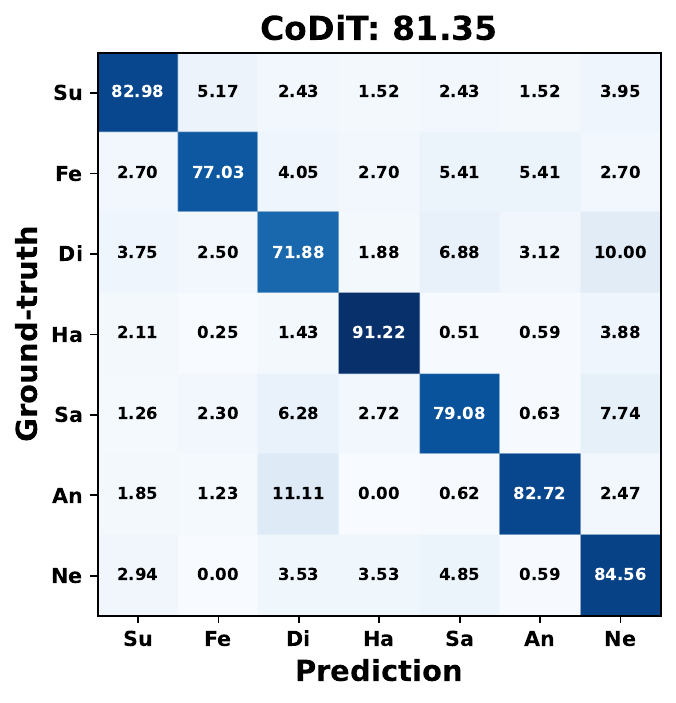}}
        \subfloat{\includegraphics[width=0.24\linewidth]{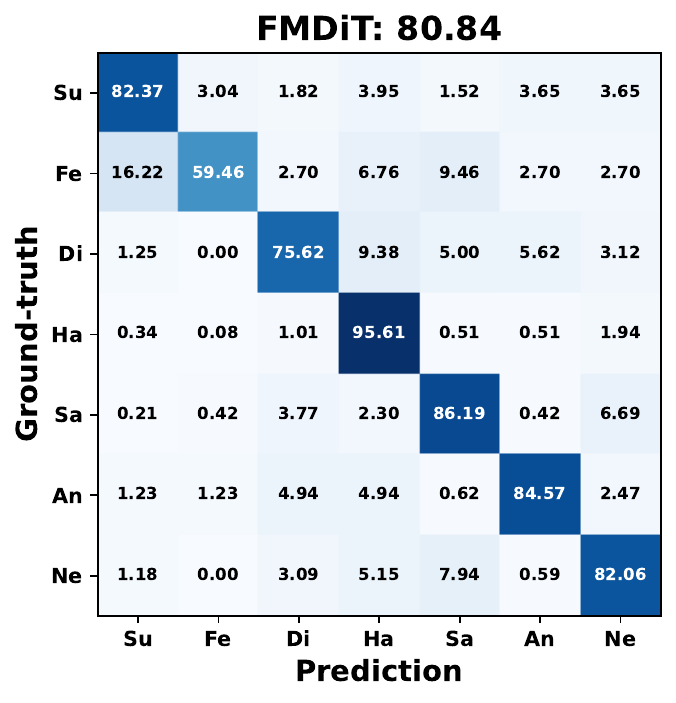}}
        \subfloat{\includegraphics[width=0.24\linewidth]{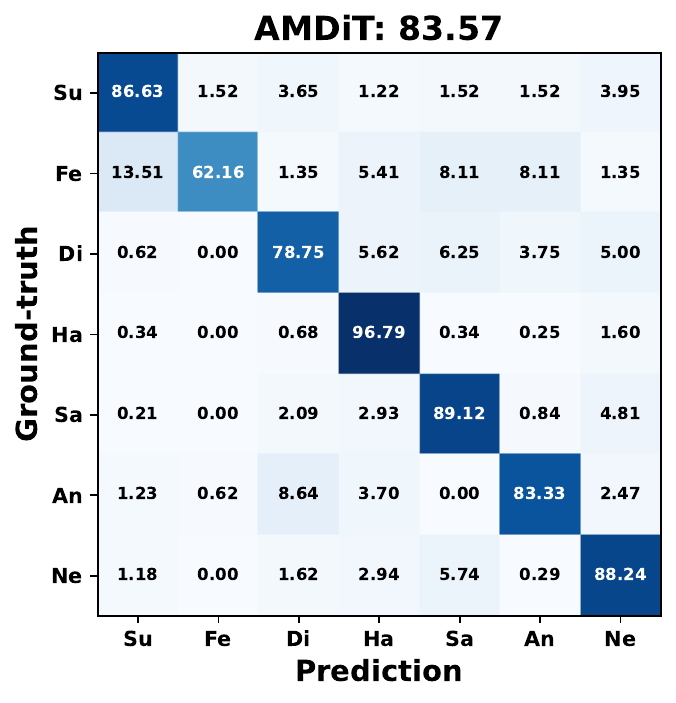}}} \\ \vspace{-0.6cm}
    \subfloat[SFEW-2.0]{\subfloat{\includegraphics[width=0.24\linewidth]{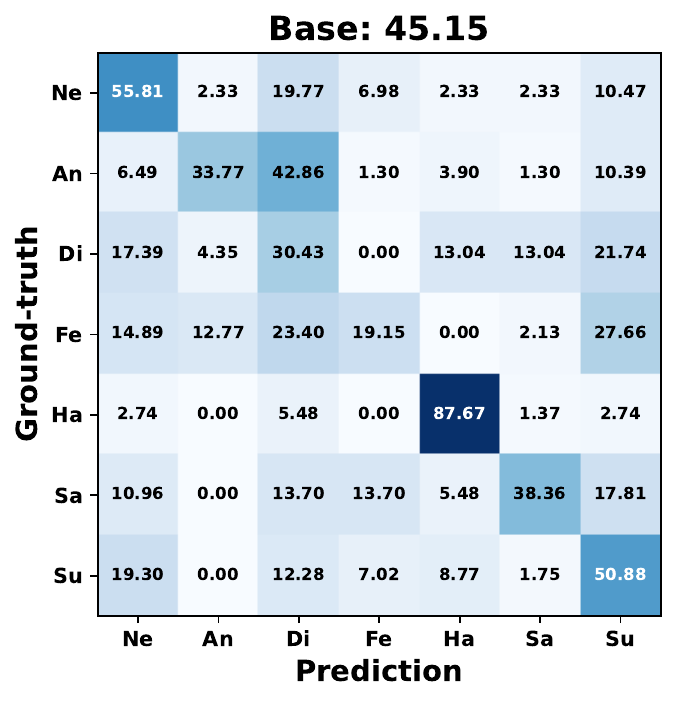}}
    \subfloat{\includegraphics[width=0.24\linewidth]{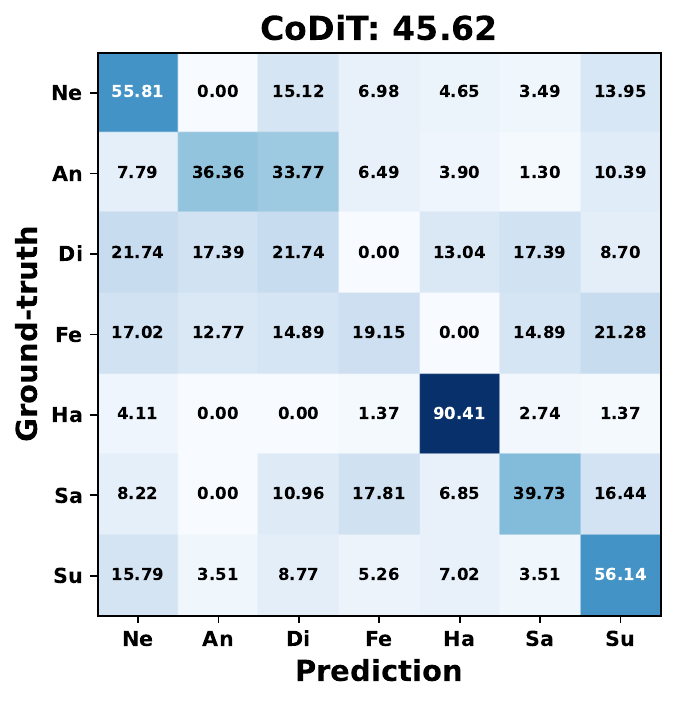}} 
    \subfloat{\includegraphics[width=0.24\linewidth]{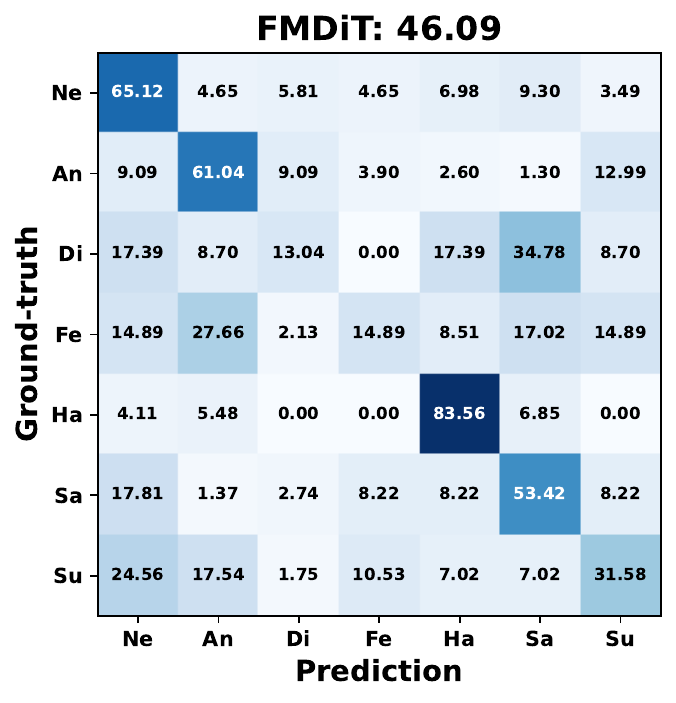}} 
    \subfloat{\includegraphics[width=0.24\linewidth]{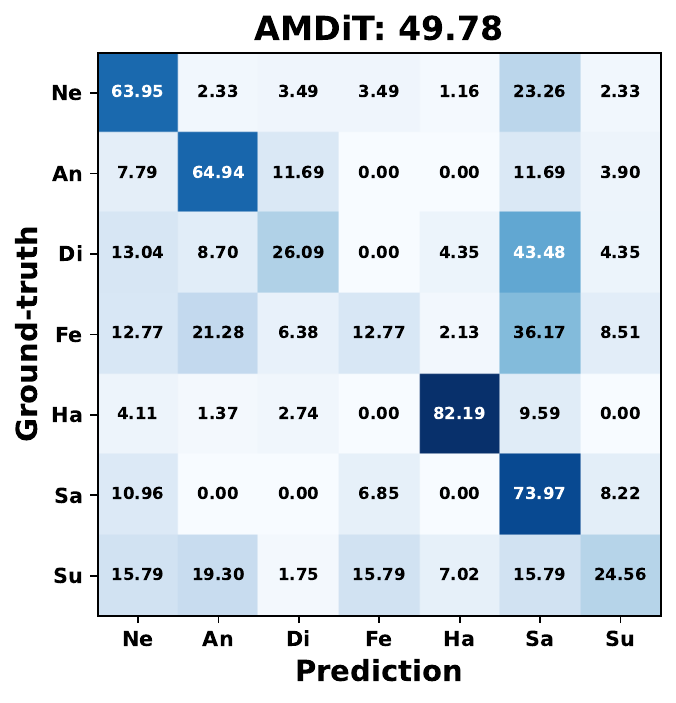}}} \\  \vspace{-0.6cm}
    \subfloat[RAF-DB\_C]{\subfloat{\includegraphics[width=0.24\linewidth]{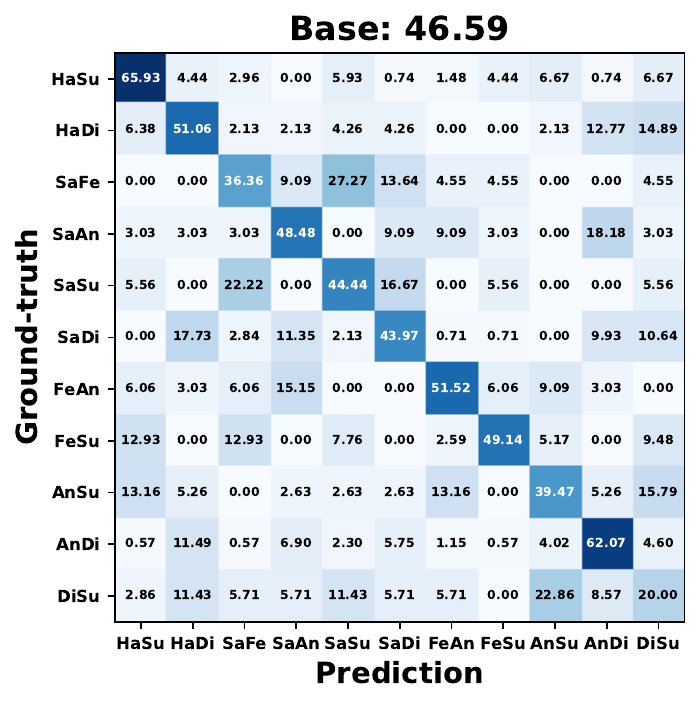}} 
    \subfloat{\includegraphics[width=0.24\linewidth]{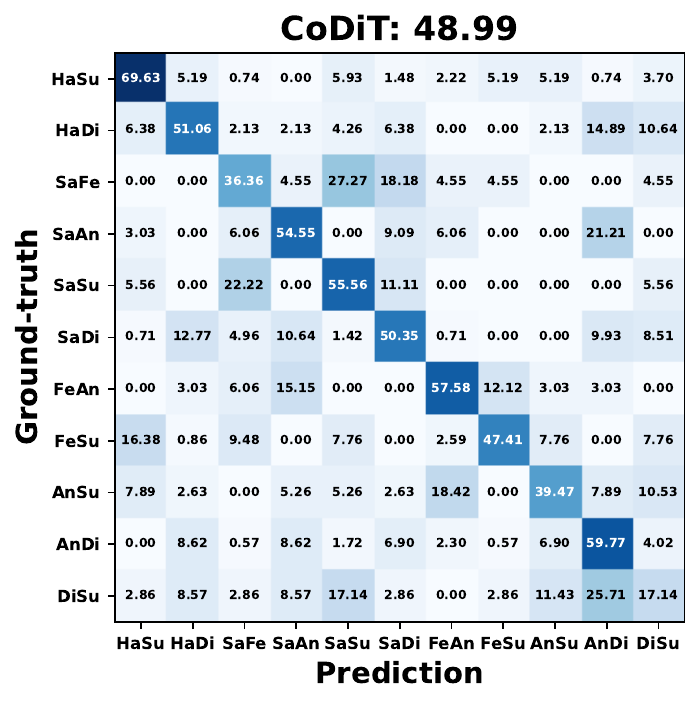}} 
    \subfloat{\includegraphics[width=0.24\linewidth]{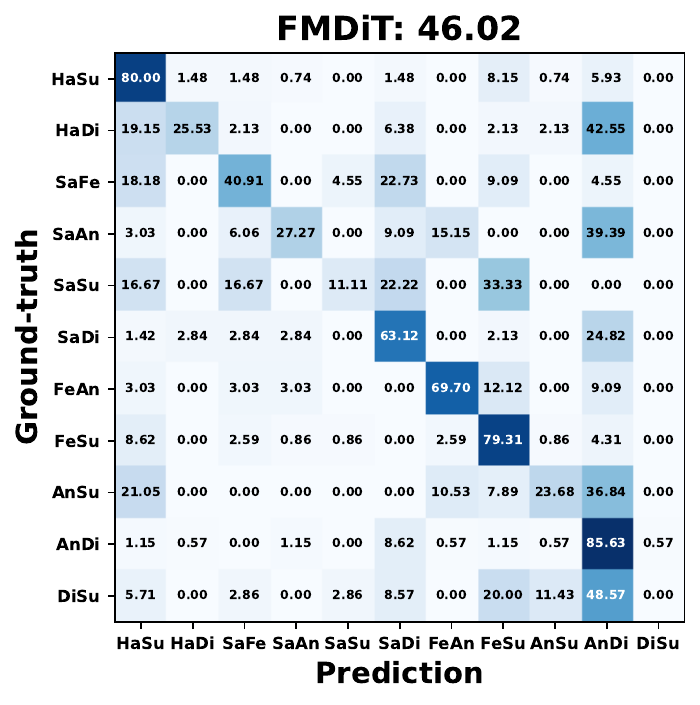}} 
    \subfloat{\includegraphics[width=0.24\linewidth]{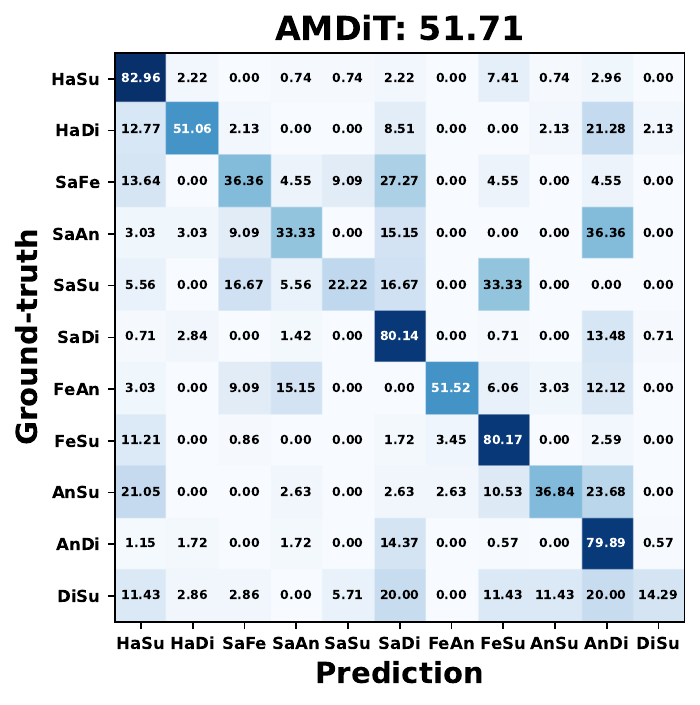}}} 
    \caption{Confusion matrices of four EmoDC variants across three datasets in 100-step evaluations. The mean class accuracy (\%) is shown above each matrix. For AMDiT, results of $c_{\rm nn}=c_{p}$ are presented.}
    \label{fig:confusion_matrix}
\end{figure*}

\subsubsection{Ablation Study}
\label{subsubsec:incremental_analysis}

To validate the effectiveness of different components in AMDiT, an ablation study is conducted across the four datasets. This evaluation compares two AMDiT variants, i.e., AMDiT ($c_{\text{nn}}=c_{\text{nc}}$), and AMDiT ($c_{\text{nn}}=c_{p}$), against four approaches: 
\begin{itemize}
    \item Zero-shot classification: The untuned Stable Diffusion model is directly evaluated for FER. Images are evaluated at a resolution of $512\times512$, consistent with its original training configuration, while other methods adopt images at the $256\times256$ resolution. 
    
    \item EmoDC Base: This baseline model trained using the vanilla tuning approach without discrepancy training (Equation (\ref{eq:base})).
    
    \item EmoDC with Contrastive noise Discrepancy Training (CoDiT): EmoDC trained with a loss function modified from $\mathcal{L}_{\rm DiT}$, written as $\mathcal{L}_{\rm CoDiT} = \mathbb{E}_{\bm{x}_{0}, t,\epsilon} \left[ ||\epsilon_{\theta}(\bm{x}_{t}, t, c_{p})-\epsilon||_{2}^{2} - \lambda_{1}||\epsilon_{\theta}(\bm{x}_{t}, t, c_{n})-\epsilon||_{2}^{2} \right]$, where $\lambda_{1}$ denotes the hyperparameter for CoDiT (see Appendix A for details).
    
    \item EmoDC with Fixed-Margin Discrepancy Training (FMDiT): EmoDC trained with a fixed margin $m=m_{f}$ in $\mathcal{L}_{\rm margin}$ (Equation~(\ref{eq:fixmargin}); hyperparameter details in Appendix A).
\end{itemize}

We compare these methods based on three incremental components: (a) whether they amplify noise-prediction errors for negative prompts, i.e., discrepancy training; (b) whether they enforce a minimal separation between the errors for positive and negative prompts, i.e., margin-based training; and (c) whether they incorporate an adaptive margin.

As shown in Table \ref{tab:incremental_results}, methods trained on facial expression datasets significantly improve FER performance compared to zero-shot classification. Discrepancy-based training methods, including CoDiT, FMDiT, and AMDiT, outperform EmoDC Base by suppressing accurate noise predictions for negative image-text pairs. Specifically, CoDiT improves accuracy over the Base model by 1.50\%, 1.65\%, 1.38\%, and 0.63\% on RAF-DB\_B, RAF-DB\_C, SFEW-2.0, and AffectNet, respectively. Further performance gains are achieved by the margin-based discrepancy training methods, i.e., FMDiT and AMDiT. For instance, on RAF-DB\_B, FMDiT yields a notable 3.55\% improvement, reaching 87.22\%. However, FMDiT struggles to adapt the margin to varying levels of noise-prediction difficulty. In contrast, AMDiT, which leverages adaptive margins, demonstrates superior performance across all datasets. Specifically, the AMDiT ($c_{\text{nn}}=c_{\text{nc}}$) variant yields accuracy improvements of 5.41\% on RAF-DB\_B, 11.62\% on RAF-DB\_C, 8.03\% on SFEW-2.0, and 4.86\% on AffectNet relative to EmoDC Base, reaching accuracies of 89.08\%, 63.51\%, 56.42\%, and 64.49\%, respectively. Notably, AMDiT ($c_{\text{nn}}=c_{p}$) achieves accuracy improvements of 6.45\% and 16.29\% on RAF-DB\_B and RAF-DB\_C, respectively, outperforming all other methods.

Given the class imbalance inherent in the datasets, both class accuracy and mean class accuracy are evaluated to ensure reliable results for real-world applications. Fig. \ref{fig:confusion_matrix} compares four EmoDC variants using 100-step evaluations. The results indicate that CoDiT improves mean class accuracy compared to the Base method. While FMDiT enhances overall accuracy, it can reduce mean class accuracy. One contributing factor to this performance drop is the challenge of using a fixed margin that is not large enough to effectively handle difficult samples in some classes, such as ``happy disgust'' in RAF-DB\_C. In contrast, the AMDiT method alleviates this issue and notably outperforms the other three EmoDC variants with a large margin in terms of mean class accuracy. This demonstrates its effectiveness in enhancing FER performance with respect to both overall and mean class accuracies.

\begin{figure*}
    \centering
    \subfloat{\includegraphics[width=0.24\linewidth]{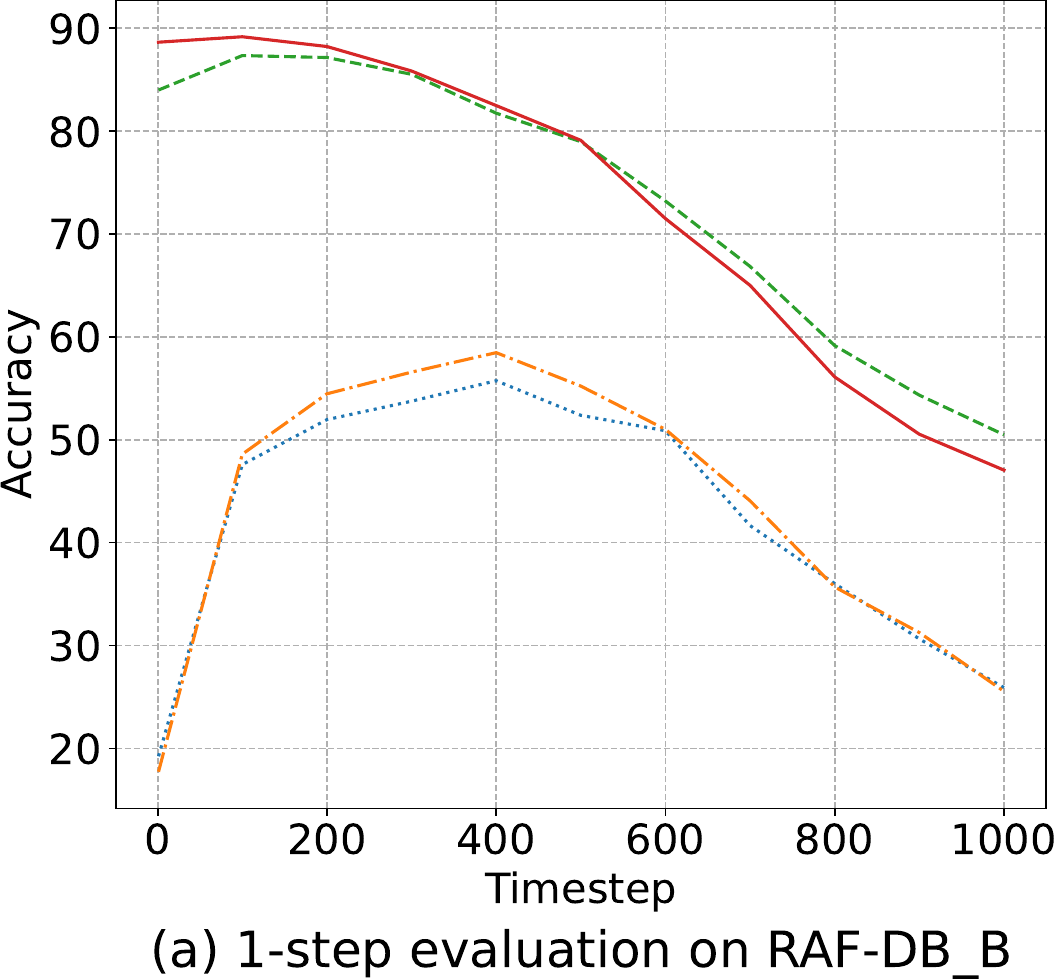}} 
    \subfloat{\includegraphics[width=0.24\linewidth]{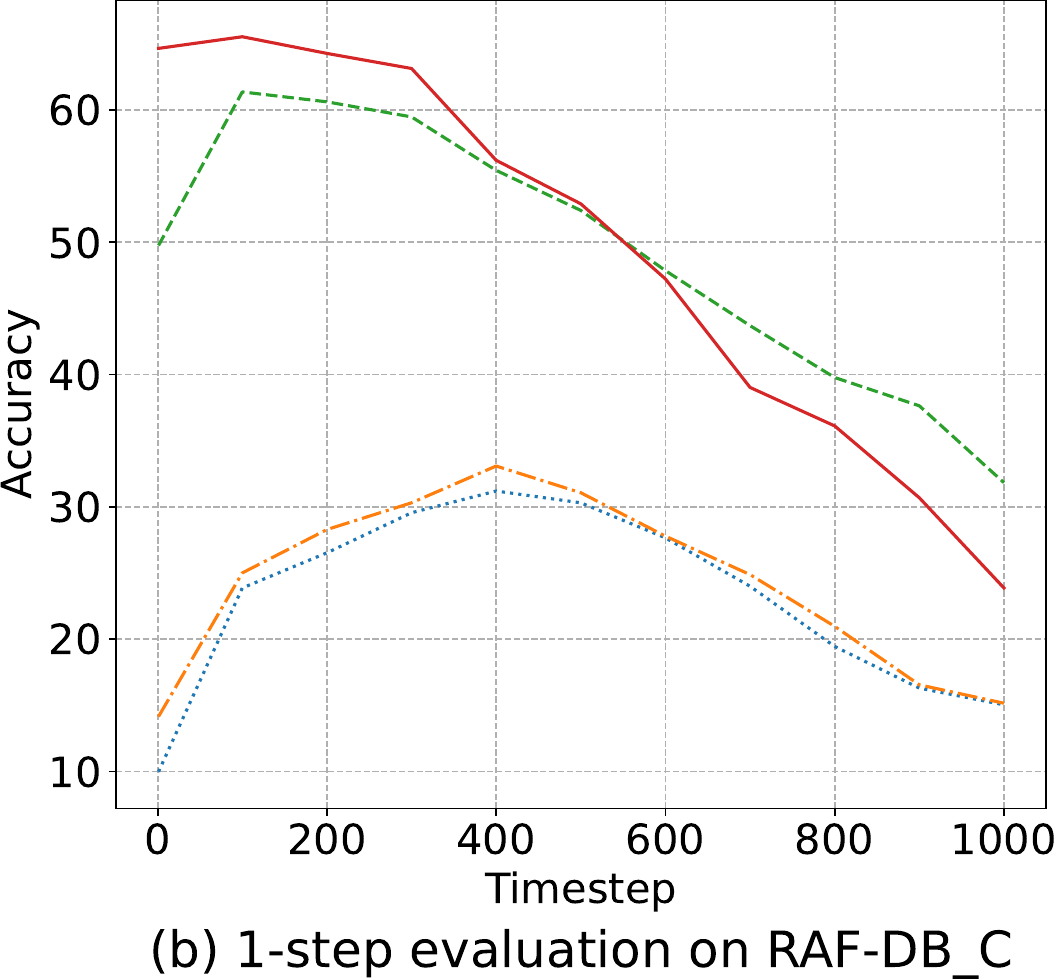}} 
    \subfloat{\includegraphics[width=0.24\linewidth]{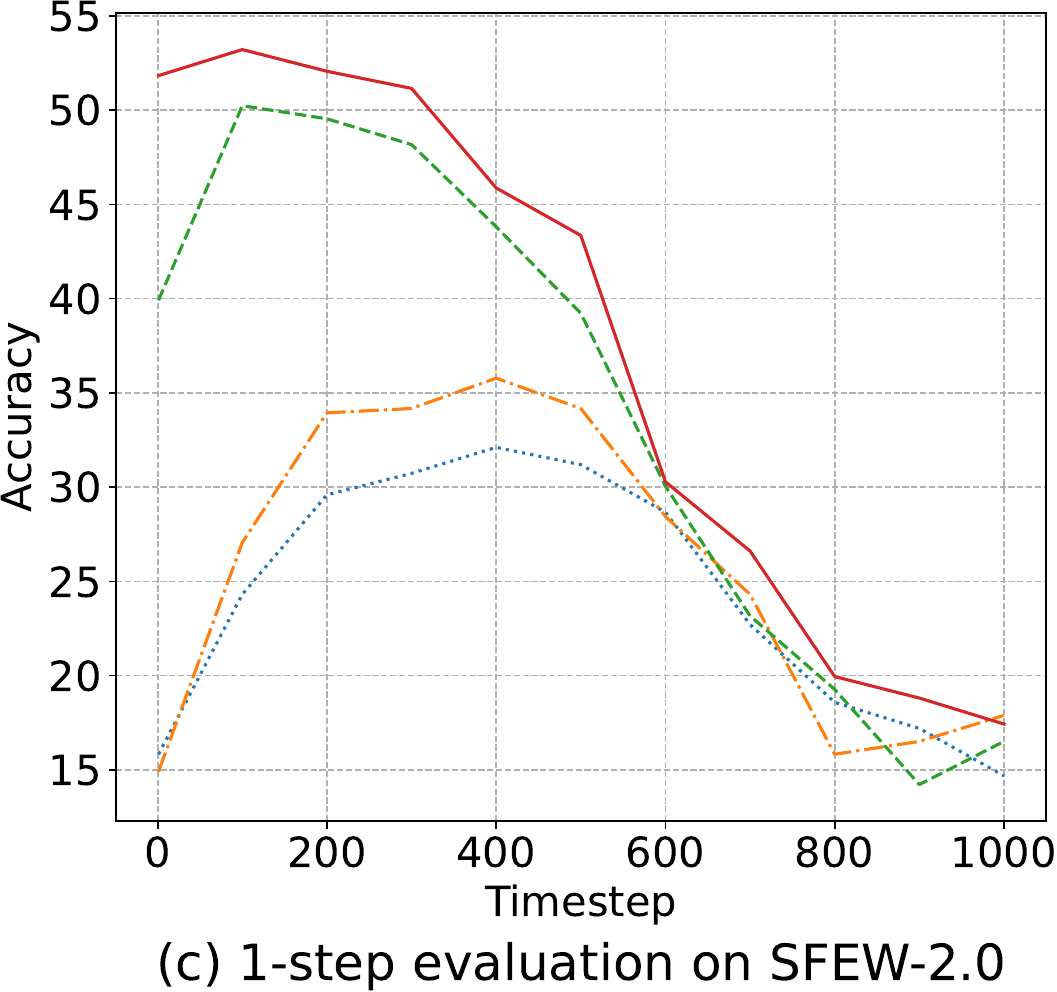}} 
    \subfloat{\includegraphics[width=0.24\linewidth]{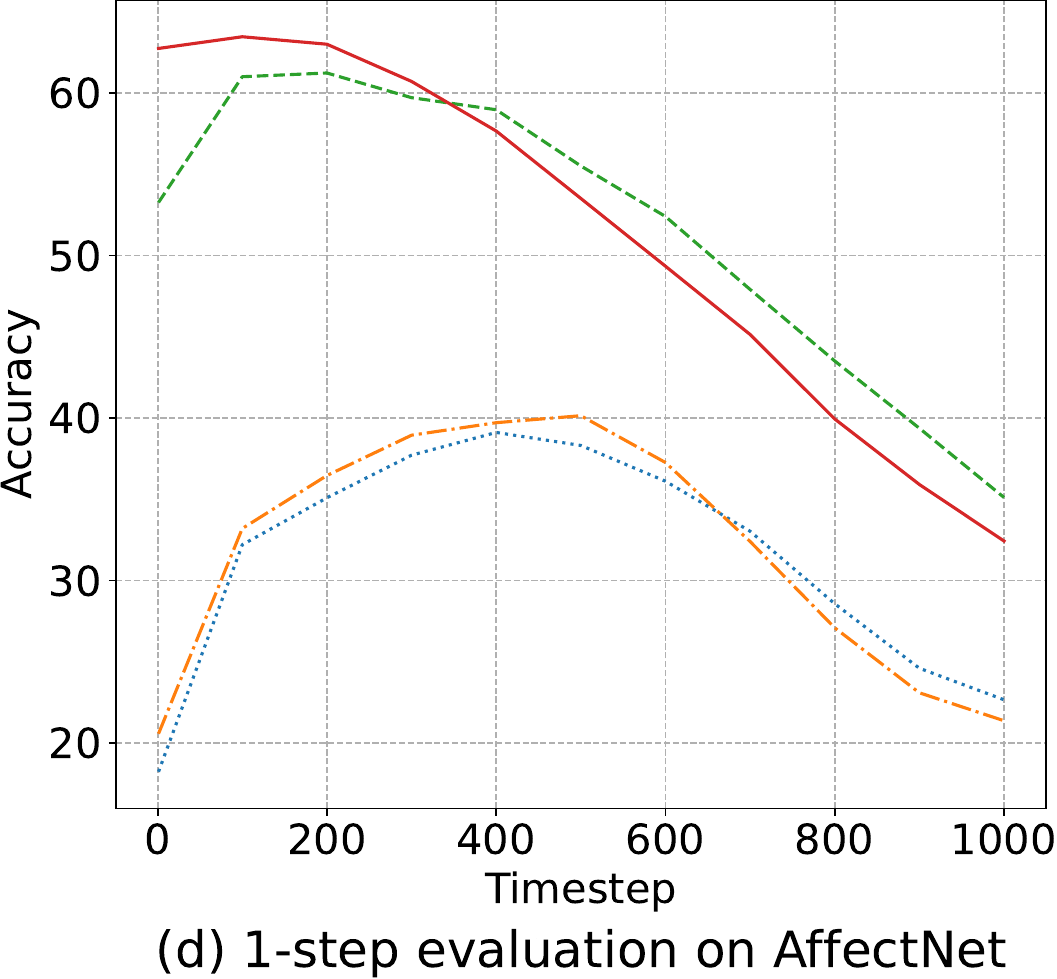}}\\
    \subfloat{\includegraphics[width=0.24\linewidth]{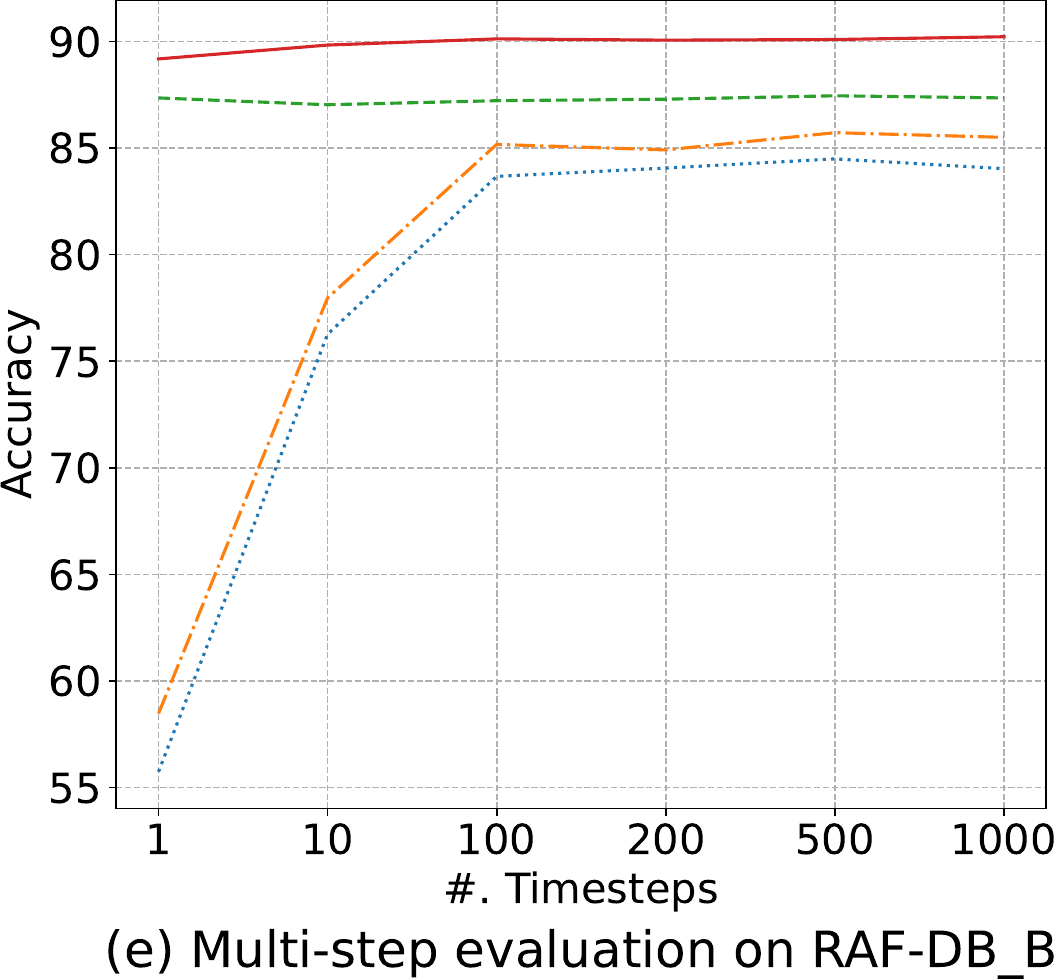}}
    \subfloat{\includegraphics[width=0.24\linewidth]{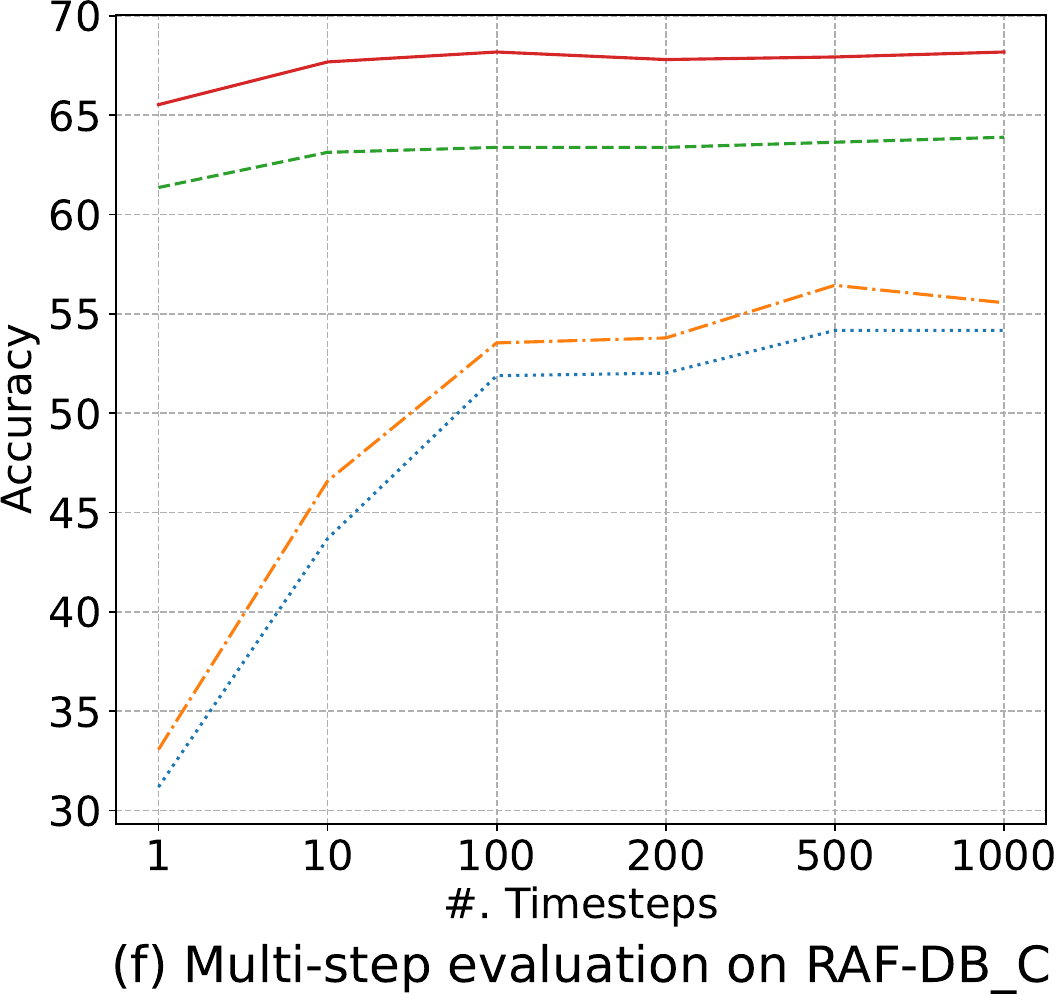}}
    \subfloat{\includegraphics[width=0.24\linewidth]{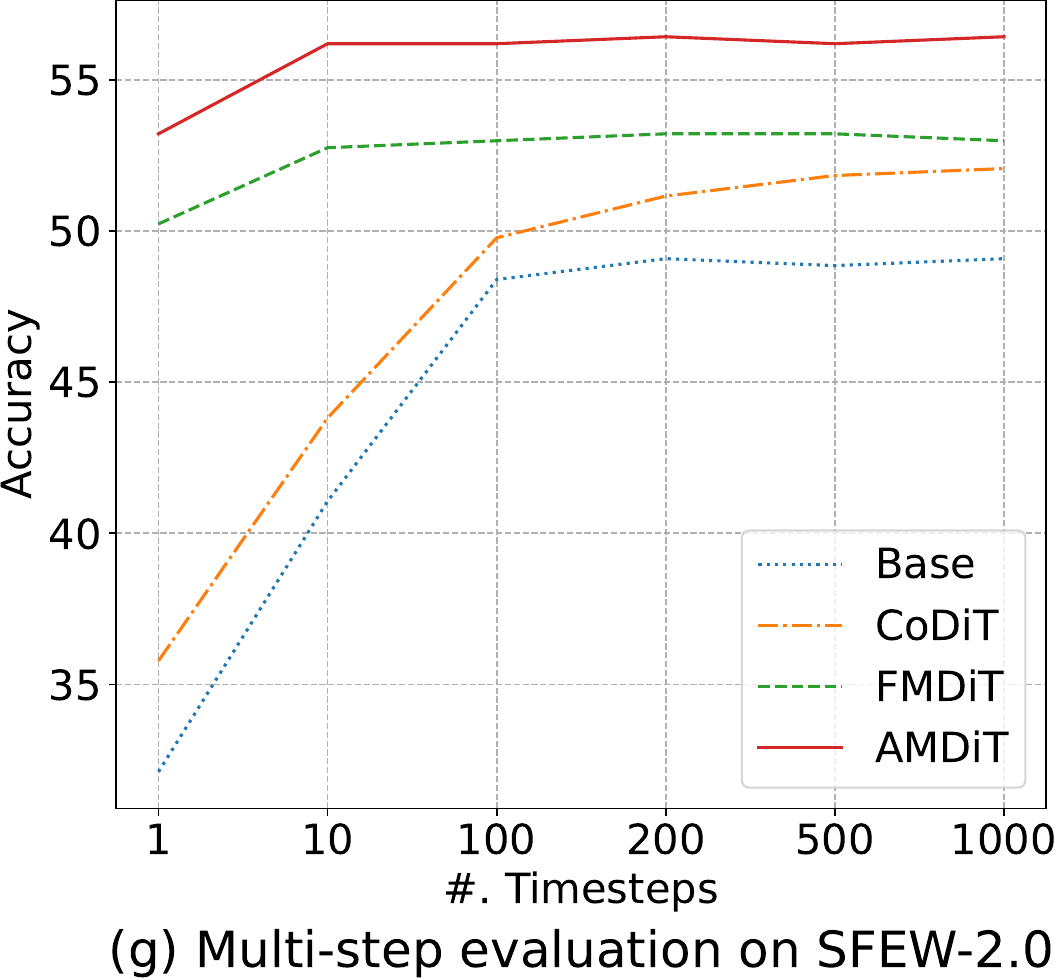}}
    \subfloat{\includegraphics[width=0.24\linewidth]{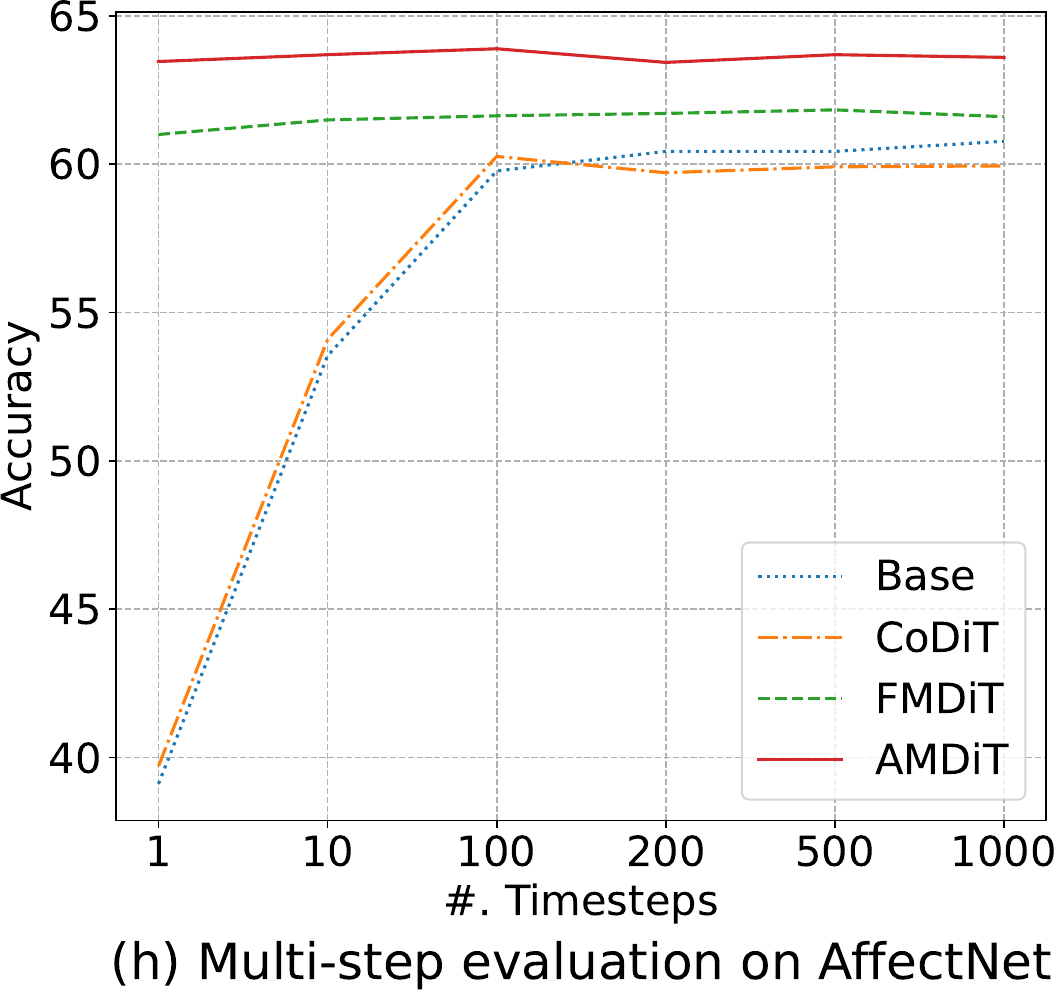}}
    \caption{1-step and multi-step evaluations of the four EmoDC variants across four datasets. For AMDiT, results with $c_{\rm nn}=c_{p}$ are presented.}
    \label{fig:1step_eval}
\end{figure*}

\subsubsection{Evaluation of Different Timesteps}
\label{subsubsec:eval_timestep}
In this section, we analyze both the 1-step and multi-step evaluations. 1-step evaluation utilizes a single timestep to obtain classification results, whereas multi-step evaluations utilize multiple evenly spaced timesteps (e.g., 10, 100, 200) to obtain the results. Evaluations were conducted on all four datasets.

\textbf{1-step evaluation.} To elucidate the efficacy and efficiency of our proposed method, we analyze the 1-step evaluation performance of EmoDC across different timesteps. The evaluation spans 11 timesteps: $\{1, 100, 200,...,900, 1000\}$. Figs. \ref{fig:1step_eval}(a)-(d) depict the 1-step evaluation results for the four EmoDC variants, leading to three key observations. First, discrepancy-based training enhances EmoDC performance compared to the baseline EmoDC Base in 1-step evaluations. While CoDiT produces only modest accuracy improvements, margin-based methods yield significant accuracy gains over both EmoDC Base and CoDiT. Second, margin-based methods alter the timestep at which EmoDC achieves peak accuracy. Specifically, both EmoDC Base and CoDiT reach their highest accuracy around the 400th timestep, consistent with findings previously reported for diffusion classifiers in \cite{li2023your}. At this timestep, EmoDC Base achieves accuracies of 55.74\% on RAF-DB\_B, 31.19\% on RAF-DB\_C, 32.11\% on SFEW-2.0, and 39.11\% on AffectNet, while CoDiT reaches 58.47\%, 33.08\%, 35.78\%, and 39.71\% on the respective datasets. In contrast, margin-based methods almost reach their highest accuracies much earlier, around the 100th timestep. At this timestep, FMDiT achieves 87.35\% on RAF-DB\_B, 61.36\% on RAF-DB\_C, 50.23\% on SFEW-2.0, and 61.00\% on AffectNet, while AMDiT further improves performance, reaching 89.18\%, 65.53\%, 53.21\%, and 63.46\% on the same datasets, respectively. Third, the peak accuracies of FMDiT and AMDiT in 1-step evaluations notably surpass those of EmoDC Base. These findings underscore the exceptional efficacy and efficiency of the proposed margin-based methods, highlighting their superior capability to optimize the single-step performance of EmoDC.

\textbf{Multi-step evaluation.} Computing mean noise-prediction errors across multiple timesteps yields higher accuracy than relying on a single timestep \cite{li2023your}. Therefore, we conduct evaluations with different numbers of timesteps to assess the potential of the four EmoDC variants. The experiments employ the 1-step evaluation and uniformly spaced 10-, 100-, 200-, 500-, and 1,000-step evaluations. For the 1-step evaluation, the 400th timestep is utilized for EmoDC Base and CoDiT, while the 100th timestep is adopted for FMDiT and AMDiT, based on results from the previous 1-step evaluation. Figs. \ref{fig:1step_eval}(e)-(h) present the multi-step evaluation results, revealing two key insights. On the one hand, discrepancy training enhances performance compared to EmoDC Base across different multi-step evaluations. Although CoDiT delivers only marginal improvements, margin-based methods, FMDiT and AMDiT, substantially outperform both EmoDC Base and CoDiT by a broad variance. This observation substantiates the significant advantage of margin-based discrepancy training in enhancing the performance of EmoDC. On the other hand, margin-based methods substantially reduce the accuracy disparity between 1-step and multi-step evaluations. Typically, the diffusion classifier requires more timesteps to achieve higher recognition accuracy \cite{li2023your}. For instance, on the RAF-DB\_B dataset, EmoDC Base needs 500 timesteps to obtain a 30\% higher accuracy over 1 timestep, but requires approximately $500\times$ the computational cost. Margin-based methods effectively mitigate this problem. While the accuracy of EmoDC Base and CoDiT stabilizes beyond 100 timesteps without significant gains, FMDiT and AMDiT stabilize using more than 10 timesteps. Moreover, the performance gap between 1-step and multi-step evaluations for FMDiT and AMDiT remains minimal. This indicates that the proposed margin-based methods can accelerate inference speed without significantly compromising accuracy by using fewer timesteps. Notably, the 1-step accuracy of the margin-based methods surpasses the peak multi-step performance of EmoDC Base and CoDiT across all four datasets. This performance gain can be attributed to the ability of margin-based methods to consistently elevate noise-prediction errors on negative pairs across different timesteps, thereby enhancing EmoDC's ability to discriminate different facial expressions. AMDiT consistently outperforms FMDiT across all datasets and multi-step evaluations, highlighting the efficacy and efficiency of the adaptive margin approach employed in EmoDC training.

\begin{table}[t]
    \centering
    \caption{Accuracy (\%) of state-of-the-art discriminative classifiers and EmoDC on RAF-DB\_B, SFEW-2.0, and AffectNet. ``FP'' indicates whether the models were pretrained on a facial image dataset before fine-tuning.}
    \label{tab:sota_basic}
    \resizebox{\columnwidth}{!}{\begin{tabular}{l|c|c|c|c}
    \hline
    \multicolumn{5}{c}{Discriminative classifiers} \\ \hline
    Methods                               & FP   & RAF-DB\_B & SFEW-2.0 & AffectNet \\ \hline
    DLP-CNN \cite{li2017reliable}         & No   & 84.13     & 51.05 & - \\
    Island loss \cite{cai2018island}      & No   & -         & 52.52 & - \\
    LBAN-IL \cite{li2021lban}             & No   & 85.89     & 55.28 & - \\
    IPA2LT (LTNet) \cite{zeng2018facial}  & No   & 86.77     & 58.29 & 57.31 \\
    RAN \cite{wang2020region}             & Yes  & 86.90     & 56.40 & 59.50 \\
    GA-FER  \cite{zhao2023geometry}       & Yes  & 87.52     & 57.11 & - \\
    DACL \cite{farzaneh2021facial}        & Yes  & 87.78     & -     & 65.20 \\
    VTFF \cite{ma2021facial}              & Yes  & 88.14     & -     & 61.85 \\
    EfficientFace \cite{zhao2021robust}   & Yes  & 88.36     & -     & 63.70 \\
    Meta-Face2Exp \cite{zeng2022face2exp} & Yes  & 88.76  & -  & 64.23 \\
    T-VPT-M \cite{dong2024text}           & No   & 89.70     & 58.49 & - \\
    APViT \cite{xue2022vision}            & Yes  & 91.98     & \textbf{61.92} & 66.91 \\
    POSTER \cite{zheng2023poster}         & Yes  & 92.05     & 57.34 & 67.31    \\
    POSTER++ \cite{mao2025poster++}       & Yes  & \textbf{92.21} & 57.57 & \textbf{67.49} \\
    \hline  \hline 
    \multicolumn{5}{c}{Emotion Diffusion Classifier (EmoDC, ours)}      \\ \hline 
    Base                                  & No     & 83.67     & 48.39  & 59.63  \\
    AMDiT ($c_{\text{nn}}=c_{\text{nc}}$)  & No     & 89.08     & \textbf{56.42} & \textbf{64.49} \\ 
    AMDiT ($c_{\text{nn}}=c_{p}$)          & No     & \textbf{90.12}      & 56.19 & 63.86 \\ \hline  
    \end{tabular}}
\end{table}

\begin{table}[t]
    \centering
    \caption{Accuracy (\%) of state-of-the-art discriminative classifiers and EmoDC on RAF-DB\_C. ``FP'' indicates whether the models were pretrained on a facial image dataset before tuning.}
    \label{tab:sota_compound}
    \begin{threeparttable}
    \begin{tabular}{l|c|c}
    \hline
    \multicolumn{3}{c}{Discriminative classifiers} \\ \hline
    Methods                               & FP     & RAF-DB\_C \\ \hline
    DLP-CNN \cite{li2017reliable}         & No     & 57.95     \\
    Separate loss \cite{li2019separate}   & No     & 58.84     \\
    ESG-Net \cite{zou2022facial}          & Yes    & 61.90     \\
    CDNet \cite{zou2022learn}             & Yes    & 63.03     \\
    Bi-center loss \cite{dong2024bi}      & Yes    & 66.16    \\
    T-VPT-M \cite{dong2024text}           & No     & 67.17      \\
    DuBEL \cite{dong2026exploring}        & Yes    & \textbf{67.68} \\
    \hline  \hline 
    \multicolumn{3}{c}{Emotion Diffusion Classifier (EmoDC, ours)}      \\ \hline 
    Base                                  & No     & 51.89     \\
    AMDiT ($c_{\text{nn}}=c_{\text{nc}}$)  & No     & 63.51  \\ 
    AMDiT ($c_{\text{nn}}=c_{p}$)          & No     & \textbf{68.18}  \\ \hline    
    \end{tabular}
    \end{threeparttable}
\end{table}

\subsubsection{Comparison to Discriminative Classifiers}

This section evaluates EmoDC trained with the proposed AMDiT method across four in-the-wild facial expression datasets. For comparison, EmoDC is benchmarked against SOTA discriminative classifiers. The results, presented in Tables \ref{tab:sota_basic} and \ref{tab:sota_compound}, also indicate whether the models were pretrained on a facial dataset prior to re-training. Pretraining on facial images typically enhances a model's ability for facial attribute analysis and improves its FER performance. 
The EmoDC Base method does not outperform any SOTA methods across the three facial expression datasets (excluding AffectNet). However, incorporating the proposed AMDiT enables EmoDC to outperform several prior SOTA methods on RAF-DB\_B and AffectNet, such as LBAN-IL \cite{ma2021facial}, RAN \cite{wang2020region}, VTFF \cite{ma2021facial}, and Meta-Face2Exp \cite{zeng2022face2exp}. Nevertheless, EmoDC still falls short of the strongest SOTA models on RAF-DB\_B and AffectNet, such as APViT \cite{xue2022vision}, POSTER \cite{zheng2023poster}, and POSTER++ \cite{mao2025poster++}, although the accuracy gap is relatively small. 
For the compound facial expression dataset RAF-DB\_C, Table \ref{tab:sota_compound} shows that EmoDC with AMDiT achieves an SOTA accuracy of 68.18\%. 
Although the proposed EmoDC with AMDiT does not yet surpass the top SOTA discriminative classifiers on basic facial expression datasets, it introduces a pioneering FER paradigm by harnessing the generative capabilities of diffusion models.

\begin{table*}[t]
    \centering
    \begin{threeparttable}
     \caption{Evaluation of EmoDCs' robustness to noise on RAF-DB\_B and SFEW-2.0 with Gaussian noise ($\mu=0$, $\sigma_{n}=\{10, 20, 30, 40, 50\}$). ``Clean'' denotes images without noise. The best results are highlighted in bold, while the second-best are underlined.}
    \label{tab:noisyacc}
    \begin{tabular}{l|c|c|c|c|c|c|c|c|c|c|c|c}
    \hline
    \multirow{2}{*}{Methods} & \multicolumn{6}{c|}{RAF-DB\_B} & \multicolumn{6}{c}{SFEW-2.0\tnote{*}} \\ \cline{2-13}
     & clean &  10   & 20    & 30    & 40    & 50  & clean &  10   & 20    & 30    & 40    & 50  \\ \hline
    APViT  \cite{xue2022vision}   & 91.98 & \textbf{90.74} & 87.71 & 84.55 & 78.94 & 70.70 & \underline{57.34} & \underline{54.82} & 45.41 & 42.43 & 33.49 & 26.15 \\
    POSTER \cite{zheng2023poster} & \underline{92.05} & \underline{90.55} & 87.16 & 79.53 & 66.23 & 46.22 & \underline{57.34} & 50.00 & 45.18 & 35.32 & 27.98 & 25.23 \\
    POSTER++ \cite{mao2025poster++} & \textbf{92.21} & 89.67 & 85.66 & 73.24 & 54.79 & 35.72 & \textbf{57.57} & 46.56 & 36.24 & 25.00 & 20.18 & 19.73   \\ 
    EmoDC Base                          & 83.67 & 79.47 & 79.04 & 76.86 & 73.99 & 70.93 & 48.39 & 49.08 & 48.62 & 46.56 & \underline{45.18} & \textbf{43.35} \\
    EmoDC AMDiT ($c_{\text{nc}}$)        & 89.08 & 88.62 & \underline{87.94} & \underline{86.51} & \underline{83.31} & \underline{80.44} & 56.42 & \textbf{55.05} & \textbf{52.98} & \textbf{49.54} & \textbf{46.33} & \underline{41.28} \\
    EmoDC AMDiT ($c_{p}$)                & 90.12  & 89.86 & \textbf{88.46} & \textbf{86.99} & \textbf{85.37} & \textbf{84.19} & 56.12 & \underline{54.82} & \underline{49.77} & \underline{46.56} & 44.27 & \underline{41.28}\\
    \hline
    \end{tabular}
    \begin{tablenotes}
        \item[*] The APViT, POSTER, and POSTER++ models were re-trained on the same aligned images as our methods.
    \end{tablenotes}
    \end{threeparttable}
\end{table*}

\begin{table*}[t]
    \centering
    \begin{threeparttable}
    \caption{Evaluation of EmoDCs' robustness to blur on RAF-DB\_B and SFEW-2.0 using Gaussian blurring with a kernel size of 15 and standard deviations $\sigma_{b}=\{2, 5, 10, 15\}$.}
    \label{tab:bluracc}
    \begin{tabular}{l|c|c|c|c|c|c|c|c|c|c}
    \hline
    \multirow{2}{*}{Methods} &  \multicolumn{5}{c|}{RAF-DB\_B} & \multicolumn{5}{c}{SFEW-2.0\tnote{*}} \\ \cline{2-11}
     & clean &  2   & 5    & 10    & 15 & clean &  2   & 5    & 10    & 15    \\ \hline 
    APViT  \cite{xue2022vision}   & 91.98 & 78.94 & 42.37 & 38.79 & 38.56 & \underline{57.34} & 52.52 & 25.92 & 24.08 & 23.39 \\ 
    POSTER \cite{zheng2023poster} & \underline{92.05} & \textbf{90.65} & \textbf{82.17} & \underline{78.68} & \underline{78.10} & \underline{57.34} & \textbf{56.65} & 49.54 & 45.64 & 44.72 \\ 
    POSTER++ \cite{mao2025poster++} & \textbf{92.21} & \underline{90.35} & 81.06 & 78.52 & 78.06 & \textbf{57.57} & 55.05 & 47.48 & 43.12 & 41.92 \\
    EmoDC Base                          & 83.67 & 61.80 & 57.46 & 60.72 & 61.34 & 48.39 & 47.02 & 48.85 & \underline{48.85} & 47.71 \\
    EmoDC AMDiT ($c_{\text{nc}}$)         & 89.08 & 88.85 & \underline{82.14} & \textbf{80.22} & \textbf{79.76} & 56.42 & \underline{56.42} & \textbf{56.88} & \textbf{56.42} & \textbf{55.96} \\
    EmoDC AMDiT ($c_{p}$) & 90.12  & 87.74 & 78.68 & 76.92 & 76.50 & 56.12 & 53.67 & \underline{50.92} & 47.94 & \underline{47.94}  \\
    \hline
    \end{tabular}
    \begin{tablenotes}
        \item[*] The APViT, POSTER, and POSTER++ models were re-trained on the same aligned images as our methods.
    \end{tablenotes}
    \end{threeparttable}
\end{table*}

\subsubsection{Robustness Evaluations}
While discriminative classifiers are vulnerable to shortcut learning and adversarial samples, generative diffusion classifiers exhibit superior robustness \cite{chen2024diffusion, chen2024robust}. In this section, we evaluate the robustness of EmoDC models against noise, blur, and adversarial attacks. All models are trained on clean images and evaluated on corrupted ones. Specifically, Gaussian noise $\epsilon_{\text{img}} \thicksim (\mu, \sigma_{n}^{2})$ is added to the original test images at varying noise levels, with $\mu=0$ and $\sigma_{n}=\{10, 20, 30, 40, 50\}$. It is worth noting that while Gaussian noise is added directly to the pixel space, Stable Diffusion inherently introduces and removes noise in the latent space \cite{rombach2022high}. Blurry images are synthesized using a Gaussian kernel with a kernel size of 15 and standard deviations $\sigma_{b}=\{2,5,10,15\}$. We also evaluate robustness against real adversarial attacks, including Square attack \cite{andriushchenko2020square}, Pixle \cite{pomponi2022pixle}, and Auto-Projected Gradient Descent (APGD) \cite{croce2020reliable}. For comparison, we benchmark EmoDCs against three open-sourced strong SOTA models: APViT \cite{xue2022vision}, POSTER \cite{zheng2023poster}, and POSTER++ \cite{mao2025poster++}.

The experimental results, presented in Tables \ref{tab:noisyacc} and \ref{tab:bluracc}, show that APViT exhibits stronger robustness to noisy samples than POSTER and POSTER++, whereas POSTER and POSTER++ excel at handling blurry facial expression images compared with APViT. 
EmoDCs exhibit a marginal performance gap compared to highly optimized discriminative models under mild corruption (e.g., blur variance $\sigma_{b}=2$ on SFEW-2.0 or RAF-DB\_B). This behavior reflects the inherent performance limitation of EmoDC on clean images. Traditional discriminative classifiers excel when corruptions are negligible.
However, although EmoDCs initially lag behind SOTA methods on clean or weakly corrupted images, they maintain relatively stable performance as noise or blur levels increase. EmoDCs deliver consistently robust performance across both corruption types. These findings highlight EmoDCs' ability to effectively mitigate performance degradation under challenging conditions, making them a practical solution for real-world facial expression recognition without requiring additional image restoration models. AMDiT-enhanced EmoDC further improves robustness, exhibiting higher resilience to both noise and blur. It surpasses all other methods when the noise level exceeds $\sigma_{n}=20$ or the blur level exceeds $\sigma_{b}=5$. These results demonstrate the effectiveness of AMDiT in enhancing EmoDC's robustness against noise and blur corruptions.

To further evaluate the robustness of EmoDC trained with AMDiT, we conduct experiments using real adversarial attacks, including Square attack \cite{andriushchenko2020square}, Pixle \cite{pomponi2022pixle}, and APGD \cite{croce2020reliable}. We implement these adversarial attacks using Torchattacks \cite{kim2020torchattacks}, adopting the following settings: 1) 50 queries for Square attack, 2) $L_2$ norm with $\epsilon_{2}=0.5$ for APGD, and 3) default Pixle settings. Table \ref{tab:result_realattack} demonstrates that SOTA discriminative classifiers, APViT \cite{xue2022vision}, POSTER \cite{zheng2023poster}, and POSTER++ \cite{mao2025poster++}, are highly vulnerable to all three adversarial attacks, suffering substantial performance degradation. In contrast, EmoDCs trained with AMDiT achieve significantly higher accuracy on the adversarial perturbed images. These results confirm the superior adversarial robustness of EmoDC when enhanced with the proposed AMDiT method.

\begin{table}
    \centering
    \caption{Evaluation of EmoDCs' robustness to adversarial attacks (Square attack \cite{andriushchenko2020square}, Pixle \cite{pomponi2022pixle}, and APGD \cite{croce2020reliable}) on RAF-DB\_B. 1-step evaluations are used for EmoDCs, with the 100th step evaluated for AMDiT models.}
    \label{tab:result_realattack}
    \begin{tabular}{l|c|c|c|c}
    \hline
    Method \textbackslash\ Attack & Clean & Square & Pixle & APGD \\ \hline
    APViT \cite{xue2022vision}  & 91.98 & 69.62  & 38.62 & 16.46 \\
    POSTER \cite{zheng2023poster} & \underline{92.05} & \underline{71.15}  & 69.78 & 33.83 \\
    POSTER++ \cite{mao2025poster++} & \textbf{92.21} & 71.09  & 72.46 & 26.11 \\
    EmoDC AMDiT ($c_{\rm nc}$) & 87.81 & \textbf{82.72} & \underline{84.35} & \underline{65.97} \\
    EmoDC AMDiT ($c_{p}$) & 89.18 & \textbf{82.72} & \textbf{87.06} & \textbf{72.33}  \\ \hline
    \end{tabular}
\end{table}

\begin{figure}
    \centering
    \includegraphics[width=\linewidth]{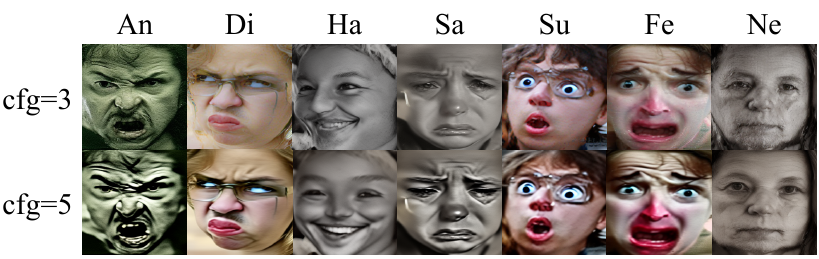}
    \caption{Examples of facial expression generation using EmoDC with AMDiT ($c_{\text{nn}}=c_{p}$), trained on RAF-DB\_B. Images are generated with classifier-free guidance scales (cfg) of 3 and 5.}
    \label{fig:ExpGen}
\end{figure}

\subsubsection{Cross-dataset evaluations}

\begin{table}[t]
    \centering
    \caption{Cross-dataset evaluation results on SFEW-2.0, FER2013Plus, and RAF-DB\_C. All models are trained on the RAF-DB\_B training set.}
    \label{tab:crossdata}
    \begin{tabular}{c|c|c|c}
    \hline
    Method & SFEW-2.0 & FER2013Plus & RAF-DB\_C  \\ \hline
    APViT \cite{xue2022vision} & 51.83 & 72.71 &  - \\
    POSTER \cite{zheng2023poster} & 44.95 & 63.52 & - \\
    POSTER++ \cite{mao2025poster++} & 48.85 & 66.50 & - \\
    EmoDC Base & 48.17 & 64.99 & 43.94 \\
    EmoDC AMDiT ($c_{\rm nc}$) & 51.72 & 77.13 & 48.48 \\
    EmoDC AMDiT ($c_p$) & \textbf{52.06} & \textbf{78.35} & \textbf{53.91} \\ \hline
    \end{tabular}
\end{table}

We conduct cross-dataset evaluations by training the models on RAF-DB\_B and testing them on three datasets: SFEW-2.0, FER2013Plus, and RAF-DB\_C. Notably, the three advanced discriminative classifiers exhibit limited cross-dataset generalization, particularly when transitioning from basic to compound facial expressions. In contrast, EmoDC demonstrates robust performance in such evaluations. This advantage stems from the fact that discriminative classifiers are constrained to predefined categories, whereas diffusion-based classifiers can generalize more effectively to unseen categories. The results, presented in Table \ref{tab:crossdata}, show that AMDiT enhances EmoDC's cross-dataset performance, outperforming both the EmoDC base model and the three strong discriminative classifiers. These findings underscore the efficacy of AMDiT in improving generalization across datasets.

\subsubsection{Generation Analysis} 
A key advantage of generative diffusion classifiers lies in their capacity to generate visual images, allowing for a clearer understanding of the knowledge acquired by the model. This capacity offers an enhancement in interpretability compared to discriminative classifiers. Diffusion models generate images through an iterative denoising process, starting from random noise sampled from a Gaussian distribution \cite{sohl2015deep, ho2020denoising, rombach2022high}. For this study, facial expression images were generated using EmoDC with AMDiT ($c_{\text{nn}}=c_{p}$). As depicted in Fig. \ref{fig:ExpGen}, the images showcase seven basic facial expressions synthesized under two classifier-free guidance scales (cfg), 3 and 5. This generative capability underscores the model's deep understanding of the underlying data distribution of facial expressions. The cfg scale controls the balance between generation diversity and fidelity \cite{ho2021classifier}: higher cfg values enhance fidelity at the expense of diversity, and vice versa. In this case, a larger cfg value results in more pronounced facial expressions characterized by exaggerated facial muscle movements. Specifically, images generated with $\text{cfg}=5$ exhibit more distinct shapes—key indicators for identifying emotional states in facial expressions. This finding aligns with \cite{jaini2024intriguing}, which highlights a human-like shape bias in diffusion classifiers. The precise generation of facial expressions arises from the accurate modeling of real data distribution by the generative diffusion model, which contributes to the high accuracy of EmoDC. Training on well-defined and diverse facial expressions enhances the model's ability to discern subtle inter-class differences (e.g., distinguishing ``surprise'' from ``fear'') and intra-class variations (e.g., degrees of ``sad''). However, EmoDC trained with AMDiT reveals two notable limitations in generated facial expressions: unnatural color tones and anomalous patterns that reduce visual coherence. Despite these limitations, the primary focus of this study remains on facial expression recognition.
Concurrent improvement of both generation and recognition performance is a direction for future research.

\section{Conclusion and Discussion}
\label{sec:conclusion}
\subsection{Conclusion}
In this paper, we explore the use of Stable Diffusion for facial expression recognition (FER) and introduce the Emotion Diffusion Classifier (EmoDC). To enhance its performance, we begin with discrepancy training, which encourages accurate noise prediction conditioned on positive class descriptions while amplifying noise-prediction errors for negative class descriptions. Building on this approach, we propose margin-based discrepancy training, which enforces a minimum margin between the noise-prediction errors for positive and negative image-text pairs. However, a fixed margin cannot adapt to sample-specific noise-prediction difficulty. To address this limitation, we develop adaptive margin discrepancy training (AMDiT), which leverages dynamic margins for training tailored to each sample. 
Experimental results demonstrate that AMDiT significantly improves accuracy across four FER datasets—RAF-DB\_B, RAF-DB\_C, SFEW-2.0, and AffectNet. AMDiT not only achieves superior accuracy but also reduces the accuracy gap between 1-step and multi-step evaluations, accelerating the inference process. Moreover, it enhances the robustness of EmoDC against noise, blur, and adversarial attacks, while simultaneously improving cross-dataset generalization to outperform state-of-the-art discriminative classifiers.

\begin{table}[t]
    \centering
    \caption{Comparison of training GPU hours and inference speed (including per-image inference time and frames per second (FPS)) for discriminative models and EmoDC variants. All evaluations were conducted on RTX 4090 GPUs.}
    \begin{tabular}{c|c|c|c}
    \hline
    Methods & GPU hours & Inference Time (ms/image) & FPS \\ \hline
    APViT \cite{xue2022vision} & 0.8 & 7.8 & 128 \\
    POSTER \cite{zheng2023poster} & 4.6 & 14.7 & 68 \\
    POSTER++ \cite{mao2025poster++} & 2.5 & 5.2 & 192 \\
    EmoDC (1-step) & 8.6 & 29.5 & 34 \\
    EmoDC (10-step) & 8.6 & 143.0 & 7 \\ \hline
    \end{tabular}
    \label{tab:computationalcost}
\end{table}

\subsection{Discussion}
While AMDiT empowers EmoDC to achieve superior robustness and cross-database generalization compared to state-of-the-art discriminative models, this improvement comes with increased computational overhead. As shown in Table \ref{tab:computationalcost}, both the training and inference times are higher than those of discriminative models, primarily due to two factors: (i) the substantial model size and (ii) the requirement for multiple model passes across different classes and timesteps. Specifically, EmoDC leverages the powerful, pre-trained Stable Diffusion model, which was originally designed for high-resolution image generation \cite{rombach2022high}. During inference, each image is encoded by the VAE encoder to generate a latent representation, which is then processed multiple times by the UNet to evaluate different categorical prompts and timesteps. These processes collectively result in slower inference. Despite these constraints, this study aims to validate the feasibility and potential of strong generative models for FER and introduces the novel AMDiT method to further improve EmoDC's performance and robustness. We acknowledge that optimizing network size and inference speed remains a critical direction for future work. Importantly, EmoDC is already viable for practical deployment. For instance, under 1-step evaluation setting, EmoDC achieves a throughput of 34 FPS, enabling real-time FER without a significant accuracy drop compared to EmoDC Base. Meanwhile, the 10-step evaluation provides a higher-accuracy alternative suitable for lower-FPS cameras or high-precision systems where speed is less critical. Furthermore, while this study focuses on static FER, extending generative classifiers to dynamic, video-based recognition using techniques such as temporal cross-attention represents a promising direction for modeling temporal emotional transitions.

\appendices
\section*{Appendix A. Hyperparameter sensitivity of EmoDC with CoDiT and FMDiT}
\begin{table}[h]
    \caption{Accuracy (\%) on RAF-DB\_B with different hyperparameter values of $\lambda_{1}$ and $m_{f}$ for EmoDC trained with CoDiT and FMDiT, respectively.}
    \label{tab:hyperparameter2}
    \centering
    \begin{tabular}{c|c|c|c|c|c}
    \hline
    Method      & Hyperparam. & 1-step & 10-step & 100-step & Avg. \\ \hline
    Base        & -           & 55.74  & 76.27   & 83.67    & 71.89 \\ \hline
    \multirow{3}{*}{CoDiT} & $\lambda_{1}=0.001$ & 56.71  & 77.71   & 84.45    & 72.96 \\
                          & $\lambda_{1}=0.005$ & 58.47  & 77.97   & 85.17    & \textbf{73.87} \\
                          & $\lambda_{1}=0.01$  & 56.71  & 77.90   & 84.45    & 73.02 \\ \hline
    \multirow{3}{*}{FMDiT} & $m_{f}=0.0001$          & 78.85  & 86.44   & 86.83    & 84.04 \\
                          & $m_{f}=0.0005$          & 87.35  & 87.03   & 87.22    & \textbf{87.20} \\
                          & $m_{f}=0.001$           & 86.96  & 86.99   & 87.19    & 87.05 \\ \hline
    \end{tabular}
\end{table}

We provide a hyperparameter sensitivity analysis of EmoDC trained with CoDiT and FMDiT, along with results from the EmoDC Base for comparisons. For the 1-step evaluation, the 400th timestep is used for EmoDCs Base and CoDiT, while the 100th timestep is used for EmoDC FMDiT (see analysis in Section \ref{subsubsec:eval_timestep}). The experimental results, presented in Table \ref{tab:hyperparameter2}, demonstrate that model performance varies with different hyperparameter settings. For CoDiT, a small $\lambda_{1}$ fails to adequately emphasize negative image-text pairs, leading to only marginal improvements over the EmoDC baseline. For FMDiT, small values of $m_{f}$ weaken EmoDC's discriminative capability due to insufficient margin separation. The optimal hyperparameter settings are empirically determined to be: $\lambda_{1}=0.005$ for CoDiT and $m_{f}=0.0005$ for FMDiT. These settings are used in the experiments in Section \ref{sec:experiments}. 

\bibliographystyle{IEEEtran}
\bibliography{reference.bib}

\vfill
\end{document}